\documentclass[final,5p,times,twocolumn]{elsarticle}

\usepackage{amssymb}
\usepackage{amsmath,amsfonts}
\usepackage{algorithmic}
\usepackage{algorithm}
\usepackage{array}
\usepackage{textcomp}
\usepackage{stfloats}
\usepackage{url}
\usepackage{verbatim}
\usepackage{graphicx}
\usepackage{amsmath}
\usepackage{color,multicol}
\usepackage{mathtools} 
\usepackage{amsfonts} 
\usepackage{array}
\usepackage{amssymb}
\usepackage{float}
\usepackage{epstopdf}
\usepackage{siunitx}
\usepackage[utf8]{inputenc}
\usepackage[english]{babel}
\usepackage{lipsum}
\usepackage{fancyhdr}
\usepackage{subcaption}
\usepackage{multirow}
\usepackage{algorithmic}

\usepackage{algorithm}

\newcommand\Tstrut{\rule{0pt}{3.6ex}}         
\newcommand\Bstrut{\rule[-2.5ex]{0pt}{0pt}}   

\newcommand{\sX}{\mathcal{X}}
\newcommand{\sY}{\mathcal{Y}}
\newcommand{\sD}{\mathcal{D}}
\newcommand{\sH}{\mathcal{H}}

\newcommand{\defas} {\overset{\underset{\mathrm{def}}{}}{=}}

\newtheorem{definition}{Definition}

\makeatletter
\def\@opargbegintheorem#1#2#3{\trivlist
   \item[]{\bfseries #1\ #2\ (#3)} \itshape}

\journal{arxiv}

\begin{document}

\begin{frontmatter}



\title{Learnability, Sample Complexity, and Hypothesis Class Complexity for Regression Models}


\author{\fnref{myfootnote} Soosan Beheshti}
\author{Mahdi Shamsi}
\address{Department of Electrical, Computer, and Biomedical Engineering, Toronto Metropolitan University,350 Victoria St., Toronto, M5B 2K3, Ontario, Canada}

\fntext[myfootnote]{Email: \tt\small{soosan@torontomu.ca}(Corresponding author)}
\begin{abstract}
The goal of a learning algorithm is to receive a training data set as input and provide a hypothesis that can generalize to all possible data points from a domain set. 
The hypothesis is chosen from hypothesis classes with potentially different complexities. Linear regression modeling is an important category of learning algorithms. 
The practical uncertainty of the target samples affects the generalization performance of the learned model. Failing to choose a proper model or hypothesis class can lead to serious issues such as underfitting or overfitting. These issues have been addressed by alternating cost functions or by utilizing cross-validation methods. These approaches can introduce new hyperparameters with their own new challenges and uncertainties or increase the computational complexity of the learning algorithm. On the other hand, the theory of probably approximately correct (PAC) aims at defining learnability based on probabilistic settings. Despite its theoretical value, PAC does not address practical learning issues on many occasions. This work is inspired by the foundation of PAC and is motivated by the existing regression learning issues. The proposed approach, denoted by $\epsilon$-Confidence Approximately Correct ($\epsilon$-CoAC), utilizes Kullback–Leibler divergence (relative entropy) and proposes a new related typical set in the set of hyperparameters to tackle the learnability issue. $\epsilon$-CoAC learnability is able to validate the learning process as a function of sample size as well as a function of the hypothesis class complexity order.
Moreover, it enables the learner to compare hypothesis classes of different complexity orders and choose among them the optimum with the minimum $\epsilon$ in the $\epsilon$-CoAC framework. Not only the $\epsilon$-CoAC learnability overcomes the issues of overfitting and underfitting, but it also shows advantages and superiority over the well-known cross-validation method in the sense of time consumption as well as in the sense of accuracy.
\end{abstract}


\end{frontmatter}
\section{Introduction}
Learning theory provides a strong framework for study and analysis of statistical inference from a data set to its underlying model \cite{vapnik1999overview, evgeniou2002regularization}. 
Supervised learning is a special case of statistical learning that deals with pairs of input data points and respective labels or target values. Supervised learning has been used in a wide range of applications, such as healthcare \cite{shailaja2018machine, qayyum2020secure, wiens2018machine}, finance \cite{culkin2017machine, dixon2020machine}, biotechnology \cite{tarca2007machine, oliveira2019biotechnology} and cybersecurity \cite{xin2018machine, handa2019machine}. In supervised learning, the goal of the learner is to estimate a mapping function, also known as hypothesis, that maps the input data to the target values, using the available training data set. The success of a hypothesis is evaluated by measuring the error between the true target vector in the training data set and the estimated target vector from the chosen hypothesis. The process of finding the function that minimizes this error is called Empirical Risk Minimization (ERM). The ability of a learning algorithm to accurately map previously unseen data, known as test data, is referred to as generalization performance. The main goal of the leaner is to find the hypothesis with the minimum possible generalization error.
Minimizing the empirical risk in the learning algorithm without any constraints can lead to a mapping function that perfectly fits all training data points, but has very poor generalization performance. One of the main reasons for this issue, also known as overfitting, especially in the regression problem, is the presence of uncertainty or, in other words, noise \cite{ghahramani2015probabilistic, vapnik1999nature}.
In this context, the No-Free-Lunch Theorem \cite{wolpert1997no, adam2019no} states that there is no universal optimization or machine learning algorithm that can successfully learn any given data set. Therefore, prior assumptions or some restrictions must be made about the relationship between the input data point and the target vector or labels, for the purpose of learning \cite{lugosi2002pattern, vapnik1999nature, shalev2014understanding}. One such assumption at the core of statistical learning theory is probabilistic modeling of the data, which assumes that all training and test data points are sampled independently and identically from a probability distribution (i.i.d.). Consequently, more training data can benefit the modeling process. Restricting the complexity of the hypothesis class can also help avoid overfitting. Applying the ERM rule in this setting leads to a probably approximately correct (PAC) hypothesis. If the generalization error of this chosen hypothesis is below an acceptable value, known as the accuracy parameter, we can declare the success of the learning algorithm. One major question in learning theory is how to choose the proper hypothesis class for a specific learning task to ensure that the learner provides a PAC hypothesis.  An example of such classes is the class of linear predictors \cite{vapnik1999nature, shalev2014understanding}. In linear regression, which is a subclass of linear predictors, the learning algorithm provides a mapping function of a given input vector, known as independent variables, to a real target vector. The hypothesis class of linear regression is a set of linear functions. Moreover, the input feature vector can be transformed to higher dimensions via a kernel function such as polynomial regression \cite{hastie2009elements}.

Due to uncertainty in the target vector, relying solely on the empirical risk minimization rule for the learning algorithm in the regression problem can lead to an overfitting problem \cite{valiant1984theory, vapnik1999nature}. Cross-validation and regularization are two well-known approaches to address this problem in the regression algorithm. The regularization method considers a constrain term for large values of parameters in the risk function, which results in penalizing the algorithm for a high variance of the parameters. The regularization method requires a regularization hyper parameter, which is chosen empirically. In addition, the method does not determine the optimal hypothesis class for the learning algorithm and, therefore, usually results in a high bias error \cite{hastie2009elements}. Cross-validation, on the other hand, resamples the training data set and divides it into a train, validation, and a test subset; then the algorithm is iterated over different models for the train subset and evaluated the generalization performance of these models over the validation subset. Therefore, in cross-validation, different models are trained from scratch on multiple subsets of the training data set, which requires more computational time and cost. Moreover, cross-validation loses a portion of data for holdout validation set and therefore for limited training data sets is not robust. However, the k-fold cross-validation method is still the most used and popular method in the machine learning literature \cite{arlot2010survey, hastie2009elements}. 

This work is motivated by two facts: recognition of the challenges and shortcomings of existing regression approaches and appreciation of the concept of PAC learnability despite its practical issues. A new learning process based on the notion of Typical sets and Kullback–Leibler divergence  \cite{cover1999elements}, is proposed. The approach, denoted by $\epsilon$-Confidence Approximately Correct ($\epsilon$-CoAC),  takes the uncertainty on the target vector into account to address the learnability of the hypothesis class.
In $\epsilon$-Confidence Approximately Correct ($\epsilon$-CoAC) learning, the notion of learnability is defined with respect to the introduced typical sets. In information theory, typical sets are defined as subsets of a given set that contain the most probable outcomes. Here, we define the typical sets in the hypothesis set, leveraging the concept of Kullback-Leibler (KL) divergence.
The approach employs this framework to define the learnability of a hypothesis class for a given training set in learning theory. We study and analyze the learnability with the necessary sample complexity for a regression model within a class of hypotheses with respect to the $\epsilon$-CoAC framework. In addition, the learnability of a hypothesis class is investigated with respect to the training data set. In this context, probabilistic bounds are derived for the generalization performance of the learning model with respect to the $\epsilon$-CoAC framework. The learning process can compare hypothesis classes of different complexity and overcome the well-known overfitting problem.
The simulation results show the superiority of the proposed method compared to the well-known cross-validation method in terms of accuracy and computation time. Unlike the cross-validation method, $\epsilon$-CoAC learnability utilizes the available data solely for the process of learning and does not require the splitting of the data. This fact improves the robustness of the learning process even in the presence of less available training data. 

\section{Notations}
Vectors are denoted by lower-case bold letters 
(for example, $\mathbf{x}$), while the $i$th element of the vectors are denoted by lower-case letters (for example $x(i)$ is the $i$-th element of $\mathbf{x}$). Moreover, upper-case letters represent random variables (for example $R$), while the samples of the random variables are denoted by lower-case letter (for example $r$).
\begin{itemize}
  \item $S=((x(1),y(1)), \dots , (x(n),y(n)))$: Training data set of size $n$, where $x(i) \in \mathcal{X}$ are input data and members of $\mathbf{x}_n = [x(1), \dots, x(n)]$ and $y(i) \in \mathcal{Y}$ are the target data and members of the target vector $\mathbf{y}_n=[y(1), \dots, y(n)]$.
  \item $\mathcal{H}_m$: Hypothesis class with order $m$.
  \item $\overline{\boldsymbol{\theta}}_{m^*}$: The true unknown parameter vector of complexity order $m^*$, where  $m^*$ is the true complexity order of the desired unknown regression model.
  \item $\widehat{\boldsymbol{\theta}}_{m,n}$:  Estimated parameter vector in hypothesis class of order $m$ from the training data set with size $n$.
  \item $\widehat{\boldsymbol{y}}_{m,n}$: Estimated target vector with length $n$ in a hypothesis class with complexity order $m$.
  \item $r^{MS}_{m, n}$: Minimum Mean Square Error (mMSE) for the hypothesis class with complexity order $m$ and training set size $n$.
  \item $r^{nrMS}_{m, n}$: Normalized Minimum Mean Square Error (nrMSE) for the hypothesis class with complexity order $m$ and training set size $n$.
  \item $r^N_{m, n}$: Desired Noise-free Mean Square Error (dNMSE) for the hypothesis class with complexity order $m$ and training set size $n$.
  \item $r^{2N}_{m, n}$: Desired Normalized Noise-free Mean Square Error (d2NMSE) Empirical risk for the hypothesis class with complexity order $m$ and training set size $n$.
\end{itemize}

\section{Learning Theory and Motivation}
Machine learning algorithm receives a training data set, denoted by $S$ in form of pairs of $(x(i),y(i))$
of length $n$:
\begin{align}
    S=((x(1),y(1)), \dots , (x(n),y(n))) \label{eqn:S}
\end{align}
The true unavailable labeling function, denoted by $f$, maps the input to its target: $\sX \rightarrow \sY$, $\forall x \in \sX$, $y = f(x)$, $y \in \sY$ where $\mathcal{X}$ and $\mathcal{Y}$ are the input domain set and output target set. The data points or instances of the domain have a probability distribution $\sD$, on $\sX$ which is denoted by $\mathcal{D}_{\mathcal{X}}$. The goal of a learner is to choose an element of a given hypothesis class $\sH$ that can map the vector of input $\mathbf{x}_n=[x(1), \dots, x(n)]$  to the output $\mathbf{y}_n=[y(1), \dots, y(n)] $ in the label set (target space). The chosen map of the learning algorithm is then used to determine the label of new data points that do not exist in the training set. For any element of the hypothesis class, $h$ , the average error between $h(x(i))$ and labels $f(x(i))=y(i)$, denoted by the true risk, is defined as
\begin{align}
    L_{\sD,f}(h) \defas \mathbb{E}_{x \sim \sD_{\mathcal{X}}}[\ell(h(x), f(x))]
    \label{eqn:TrueRisk}
\end{align}
where $\mathbb{E}$ denotes expectation and $\ell$ is a desired loss function that determines the cost of using hypothesis $h$ to label the domain set instead of the true labeling function $f$. As neither the probability distribution $\sD_{\mathcal{X}}$ nor the true labeling function $f$ is known, this true risk in (\ref{eqn:TrueRisk}) cannot be calculated directly. 
However, if the samples are good representations of their underlying probability, the true risk is estimated by the observed sample as follows (\textbf{empirical risk}):
\begin{align}
    L_S(h) &= \frac{1}{n} \sum_{i=1}^{n} \ell(h(x(i)),f(x(i))) \label{eqn:ermloss}
\end{align}
where $y(i) = f(x(i))$. To find the optimum $h \in \sH$, the learning algorithm uses \textbf{empirical risk minimization (ERM)} rule to determine the optimal hypothesis denoted by $h_S$ 
\begin{align}
    h_S = \underset{h \in \mathcal{H}}{\arg \min} \; L_S(h) \label{eqn:erm}
\end{align}
Note that $L_{\sD,f}(h_S)$ in (\ref{eqn:TrueRisk}) denotes the unavailable true risk of the hypothesis chosen by the ERM minimization. In the following section, we briefly explain Probably Approximately Correct (PAC) learning and discuss the motivation of this work.
\subsection{Probably Approximately Correct (PAC) Hypothesis Class}

PAC learning is a framework for analyzing and evaluating the performance of a learner. It provides a mathematical approach to determine the accuracy and confidence of a machine learning algorithm. Through this framework, one can measure the number of training examples required to reach a specific level of accuracy and evaluate the performance of the learning algorithm for generalization. 
PAC learning is concerned with the learnability of a hypothesis class $\mathcal{H}$ with respect to any desired mapping function $f \in \mathcal{H}$ for any probability distribution $\mathcal{D}_{\mathcal{X}}$ on the domain set. Learnability is defined based on accuracy and confidence probabilistic parameters. In this scenario, the output hypothesis of the learning algorithm is a given mapping that provides one $h_S$ using the training data set. One example of such mapping is the result of the empirical risk in (\ref{eqn:ermloss}).

\textit{PAC Learnable Hypothesis Class}: Consider a desired accuracy parameter $\epsilon$ and confidence parameter $\delta$, hypothesis class $\mathcal{H}$, and a learning algorithm that chooses $h_S$ based on the empirical risk minimization. The hypothesis class $\mathcal{H}$ is PAC learnable if the following holds
\begin{align}
    L_{\sD, f}(h_S) \leq \epsilon
\end{align}
 with probability $1-\delta$ for any mapping functions $f\in \mathcal{H}$ and any probability distributions $\mathcal{D}_{\mathcal{X}}$. Equivalently, $1-\delta$ is the probabilistic confidence in the output hypothesis of the learner. The minimum value $n_{(\epsilon, \delta)}$ that satisfies this condition is defined as the sample complexity \cite{valiant1984theory}.
PAC learning addresses the relation between sample complexity and generalization ability of the learning models. For example, it has been used to derive probabilistic bounds on the sample complexity of finite hypothesis classes, half-spaces against the uniform distribution \cite{long1995sample} and neural networks \cite{golowich2018size}.

\subsection{Uncertainty Issues in the Learning process and Motivation}
In this section, we discuss three issues of learning in practical applications. The first issue concerns the uncertainty in the input feature vector that we briefly discuss here. The second and third issues are regarding the uncertainty in the target vector as well as uncertainty in the richness of the hypothesis class, which are the motivation of our work.

\textit{Uncertainty in the Input Feature Vector:}
 PAC learning algorithm relies on an assumption that the input feature vector $\mathbf{x}$ is generated from the domain set $\mathcal{X}$ with respect to the probability distribution $\mathcal{D}_{\mathcal{X}}$. However, in practical applications, usually partial or no prior information is available on $\mathcal{D}_{\mathcal{X}}$ \cite{wang2022generalizing}. The problem of uncertainty in the feature vector has been addressed in the literature and deals with issues such as generalization of the output hypothesis once a new data set is available or concerns with noisy input \cite{farahani2021brief, gretton2009covariate, wen2014robust, zhou2022domain, ghifary2016scatter, mahajan2021domain}. Another related issue does not deal with the choice of possible distributions, but with the structure of the observed data and whether the observed data are representative of the desired underlying probability distribution \cite{mehrabi2021survey, zafar2019fairness}. Although this issue is indirectly related to the finiteness of the available sample, it can cause serious problems in the learning process. An example of this issue is in \cite{chamon2020empirical} where the goal is to predict whether an individual salary is more than a certain amount based on features such as age, weekly hours, marital status, and education. It is discussed that since the number of male samples in the available data set is greater than the number of females, the learning model can be heavily biased towards men.

\begin{figure}[htb]
\centering
\includegraphics[trim=20cm 0cm 20cm 0cm, scale=1]{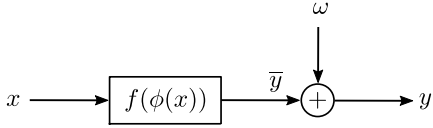}
\caption{Example of a noisy target vector.}
\label{fig:Control2}
\end{figure}

\textit{Uncertainty in the Target Vector and Uncertainty in the Hypothesis class:}
Another issue in the learning process is that of uncertainty of the target vector. 
A possible scenario is shown in Figure \ref{fig:Control2}, where the resulted target vector $f(\phi(\mathbf{x}))$ can be corrupted by an additional noise $\omega$. This issue also plays an important role in the quality and generalization performance of the learning algorithm. \cite{song2022learning} provides a comprehensive survey of methods that address the problem of noisy labeling in deep neural networks. There are many recent researches in this area and they mostly concentrate on either presenting a new validation method for the learning process or modifying the empirical risk in (\ref{eqn:erm}) by, for example, adding some correction terms \cite{ma2018dimensionality,zhang2021learning,han2018masking, wang2019symmetric,wong2016constrained,hastie2009elements}.
 
The third interesting and important related issue is that the PAC learning algorithm relies on the true labeling function $f(\phi(\mathbf{x}))$ being a member of the hypothesis class. However, in practical applications, this is not a valid assumption. In fact, as it is not known whether the labeling function belongs to the hypothesis class, there are many research works on
comparing competing hypothesis classes in the learning process. Consequently, different methods define choose an optimum hypothesis class based on their define cost function, and notions of hypothesis class selection and model selection are often used interchangeably. Approaches such as cross-validation or modification of the risk function by adding additional terms are examples of these methods \cite{kim2009estimating, cherkassky2003comparison, fushiki2011estimation}.  

 This work is motivated by existing challenges related to the uncertainty of the target vector and the uncertainty of the hypothesis class in regression modeling. These issues are known to cause serious problems, such as overfitting or underfitting \cite{fushiki2011estimation, cherkassky2003comparison}, or a high time consumption in cross-validation approaches. Inspired by the useful fundamentals of PAC learning and concerned with practical aspects of the learning process, our focus will be on the learnability of a regression model with uncertainty on the output vector due to the additive noise and uncertainty on the hypothesis class complexity. We examine the role of data length and hypothesis complexity in the context of probabilistic confidence.  

\section{Regression Algorithm with Uncertainty on the Target Vector}
\label{sec:LearningTheory}
In this section, we first briefly review basic notations of the regression modeling and illustrate the effect of output uncertainty in the learning process.  Consider a noise-free target vector of length $n$ denoted by $\overline{\mathbf{y}}_n=[\overline{y}(1), \dots,\overline{y}(n)]^T$, $\bar y(i) \in \mathbb{R}$, that is generated by the linear mapping function $f$ as follows:
\begin{align}
    \overline{\mathbf{y}}_n &= f(\phi(\mathbf{x}_n)) = \mathbf{A} \overline{\boldsymbol{\theta}} \label{eqn:ybar}
\end{align}
where $\overline{\boldsymbol{\theta}} = [\bar \theta(1), \dots, \bar \theta(m)]^T$ is the true unknown parameter vector and the elements of the matrix $\mathbf{A}=\phi(\mathbf{x}_n)$ are functions of the available input vector $\mathbf{x}_n = [x(1), \dots, x(n)]^T$ of length $n$. Depending on the application, these elements can be a linear function of $x$, such as linear time invariant modeling, or a nonlinear function of $x$ such as kernel functions. For example, in $m$-th order polynomial regression, the kernel functions of the $i$-th row of matrix $\mathbf{A}$ are $\mathbf{A}(i,:) = [1, x(i), x^2(i), \dots, x^m(i)]$ where $x(i) \in \mathbf{x}_n$.
The available noisy target vector is:
\begin{align}
    \mathbf{y}_n &=  \mathbf{\overline{y}}_n + \boldsymbol{\omega} = \mathbf{A} \overline{\boldsymbol{\theta}} + \boldsymbol{\omega} \label{eqn:noisyoutput}
\end{align}
where $\boldsymbol{\omega}$ is a sample of independent identically distributed (i.i.d) Gaussian zero mean distribution $ \mathcal{D}_\omega \sim \mathcal{N}(0, \sigma^2_\omega)$. The regression learning algorithm estimates the parameter vector $\overline{\boldsymbol{\theta}}$ from the available training data. Elements of the hypothesis class are now equivalently parameters in the form of $\boldsymbol{\theta}$:
\begin{align}
    h \leftrightarrow  \boldsymbol{\theta}
\end{align}
and the output of each hypothesis $h$ given the input data, is in the form of
\begin{align}
    \boldsymbol{\mu}_{\boldsymbol{\theta}}=h(\mathbf{x}_n)=\mathbf{A} \boldsymbol{\theta} \label{eqnn:muAndy}
\end{align}
Minimizing the empirical loss in (\ref{eqn:ermloss}) provides the optimum hypothesis ${h}_S$ (equivalently $\widehat{\boldsymbol{\theta}}$) and provides an estimate for the target vector:
\begin{align}
    \widehat{\mathbf{y}}_n =  h_S(\mathbf{x}_n) = \mathbf{A} \widehat{\boldsymbol{\theta}}  = \boldsymbol{\mu}_{\widehat{\boldsymbol{\theta}}} \label{eqn:firstyhat}
\end{align}
Regression algorithms have proposed different loss functions $\ell$, and different optimization approaches to solve this problem \cite{ranstam2018lasso, marquardt1975ridge, friedman2010regularization}. 
The well-known and most used loss function in this scenario is the variance of error in the output estimation. The corresponding empirical risk in (\ref{eqn:ermloss}) in this case is the \textbf{Mean Square Error (MSE) empirical risk}, which is the $\ell_2$ norm of the output error. For elements of the hypothesis class, this MSE is:
\begin{align}
     L_S(\boldsymbol{\theta}) = r^{MS} (\boldsymbol{\theta})= \frac{1}{n} \sum_{i=1}^{n} ({\mu}_{\boldsymbol{\theta}}(i) -y(i))^2 \label{eqn:lerSample}
 \end{align}
Minimizing the MSE empirical risk in (\ref{eqn:lerSample}) leads to the following ordinary least squares solution:
\begin{align}
    \widehat{\boldsymbol{\theta}} = \underset{\boldsymbol{\theta} \in \mathcal{H}}{\arg \min} \; r^{MS}(\boldsymbol{\theta}) =(\mathbf{A}^T\mathbf{A})^{-1}\mathbf{A}^T\mathbf{y}_n
    \label{eqn:OLS22}
\end{align}
To calculate this estimate, it is assumed that the input is rich enough such that $\mathbf{A}^T\mathbf{A}$ is a full rank and invertible matrix. The minimum empirical risk $r^{MS}$ in (\ref{eqn:erm}), denoted by \textbf{minimum Meas Squared Error (mMSE)} is
\begin{align}
    r^{MS} (\widehat{\boldsymbol{\theta}}) = \frac{1}{n} \sum_{i=1}^{n} (\widehat{y}(i) -y(i))^2 = \frac{1}{n}\left \|\widehat{\mathbf{y}}_n - {\mathbf{y}}_n  \right \|_2^2 \label{eqn:rm}
\end{align}
Methods such as \cite{kim2009estimating, cherkassky2003comparison, fushiki2011estimation, awad2015support} modify the MSE empirical risk by adding additional terms. The goal of these modifications is to control the number and values of parameters in a given hypothesis class to avoid overfitting. 
There are other methods that, due to the existence of outliers or uncertainty in the input vector, move to different loss functions from the MSE \cite{balasundaram2020robust}. In this paper, we do not consider issues related to the input of the regression model, such as outliers or uncertainty in the input. We use the MSE loss function for the parameter estimation and consider issues such as an unknown number of parameters and overfitting in the learning algorithm. However, in the following sections, we illustrate that these issues can be properly addressed by not diverting from the MSE loss function for parameter estimation but by comparing different related cost functions for complexity and validity analysis of the estimators of different order. 
\section{$\epsilon$-Confidence Approximately Correct ($\epsilon$-CoAC) Learnability}

\begin{figure}[htb]
\centering
\includegraphics[trim=20cm 0cm 20cm 0cm, scale=1]{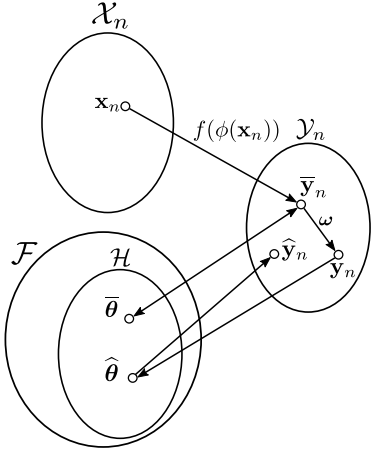}
\caption{Important elements of the regression learning process.}
\label{fig:CoAC}
\end{figure}

Figure \ref{fig:CoAC} illustrates the important elements of the learning process. In this figure we assume that the true parameter $\overline{\boldsymbol{\theta}}$ is a member of the hypothesis class (this assumption is later relaxed). 
The sequence of parameter estimation is as follows: The observed input $\mathbf{x}_n$ generates the unobserved noise-free target vector $\overline{\mathbf{y}}_n$ which is then corrupted with the additive noise $\boldsymbol{\omega}$. The noise-free target data $\overline{\mathbf{y}}_n$ is generated by the desired true mapping function $f$ which in the case of linear regression is equivalent to parameter $\overline{\boldsymbol{\theta}}$ in the parametric hypothesis class $\mathcal{H}$ in (\ref{eqn:ybar}).  
The available noisy target vector $\mathbf{y}_n$ is used to estimate the desired $\overline{\boldsymbol{\theta}}$ which is the least square solution $\widehat{\boldsymbol{\theta}}$ from (\ref{eqn:OLS22}). The estimated $\widehat{\boldsymbol{\theta}}$ itself generates an estimate of the noise-free target through (\ref{eqn:firstyhat}) denoted by $\widehat{\mathbf{y}}_n$. 

\subsection{Typical Set in Regression Learning}
Kullback-Leibler, which is also known as the relative entropy, indicates the disparity, or, in other words, the divergence between the true underlying distribution, $P$, of the data and the estimated distribution $Q$, and is denoted as follows:
\begin{align}
    D_{KL}(P \parallel Q) =\int _{-\infty }^{\infty }p(x)\log \left({\frac {p(x)}{q(x)}}\right)\,dx \label{eqn:KLOriginal}
\end{align}

Inspired by the role of the typical sets in source coding and motivated by the notion of Kullback–Leibler divergence in (\ref{eqn:KLOriginal}) for the notion of the distance between probability distributions, we define Typical sets for Regression Learning as follows.

\begin{definition}[$\epsilon$-Typical Set]
$\epsilon$-Typical set of a desired distribution $P$ is a set of distributions that satisfy the following.
\begin{align}
    T_{\epsilon}(P) = \left \{ Q : D_{KL}\left(P || {Q}\right) < \epsilon \right \} \label{eqn:kldiv}
\end{align} 
\end{definition}
\subsection{$\epsilon$-Confidence Approximately Correct ($\epsilon$-CoAC) Hypothesis Set}
Each hypothesis in the hypothesis class, $\forall \boldsymbol{\theta} \in \mathcal{H}$, generates elements of the noisy target space with a linear regression model corrupted with a Gaussian distribution with variance $\sigma_\omega^2$:  
\begin{align}
    \boldsymbol{\theta}: \mathcal{D}_{\mathbf{y}}(\boldsymbol{\theta}) &\sim \mathcal{N}(\boldsymbol{\mu}_{\boldsymbol{\theta}}, \sigma^2_\omega) \label{eqn:klmutheta}
\end{align}
where the mean of the distribution represents the noise-free element denoted by $\boldsymbol{\mu}_{\boldsymbol{\theta}}$:
\begin{align}
     \boldsymbol{\mu}_{\boldsymbol{\theta}} &= \mathbf{A}\boldsymbol{\theta} \label{eqn:muteta}
\end{align}
For the probability distribution generated by the true parameter $\overline{\boldsymbol{\theta}
}$, the mean is $\overline{\mathbf{y}}_n$ itself. 
\begin{align}
\overline{\boldsymbol{\theta}}:\mathcal{D}_{\mathbf{y}}(\overline{\boldsymbol{\theta}}) &\sim \mathcal{N}(\overline{\mathbf{y}}_n, \sigma^2_\omega) \label{eqn:klybartheta}
\end{align}
 where
\begin{align}
    \overline{\mathbf{y}}_n = \boldsymbol{\mu}_{\overline{\boldsymbol{\theta}}} =\mathbf{A}\overline{\boldsymbol{\theta}}
\end{align}

\begin{definition}[$\epsilon$-Confidence Approximately Correct ($\epsilon$-CoAC) Hypothesis Set]
A set of hypotheses, which is a subset of the considered Hypothesis class $\mathcal{H}$, is $\epsilon$-Confidence Approximately Correct with respect to the true parameter $\overline{\boldsymbol{\theta}}$ if they are a member of the Typical set of $\mathcal{D}_{\mathbf{y}}(\overline{\boldsymbol{\theta}})$:
\begin{align}
    T_{\epsilon}(\overline{\boldsymbol{\theta}}, \mathcal{H}) = \left \{ \boldsymbol{\theta} \in \mathcal{H} : D_{KL} \left ( \mathcal{D}_{\mathbf{y}}(\overline{\boldsymbol{\theta}}) \parallel \mathcal{D}_{\mathbf{y}}(\boldsymbol{\theta}) \right ) < \epsilon \right \} \label{eqn:COACOriginal}
\end{align} 
\end{definition}
Figure \ref{fig:MSE1} shows an example of a $\epsilon$-CoAC hypothesis set.

\subsection{$\epsilon$-CoAC Target Set and $\epsilon$-CoAC Learnability}

Considering the fact that the distributions imposed by the hypothesis class in this setting are Gaussian with the same variance and different means in (\ref{eqn:klmutheta}) and (\ref{eqn:klybartheta}), the Kullback-Leibler distance between the true hypothesis and a hypothesis in the hypothesis class $\mathcal{H}_m$ is equivalent to the corresponding $\ell_2$ distance of the true target vector $\overline{\mathbf{y}}_n$ and $\boldsymbol{\mu}_{\boldsymbol{\theta}}$ (\ref{eqn:muteta}):  
\begin{align}
    D_{KL} \left ( \mathcal{D}_{\mathbf{y}}(\overline{\boldsymbol{\theta}}) \parallel \mathcal{D}_{\mathbf{y}}(\boldsymbol{\theta}) \right ) = \frac{1}{2}(\frac{|| \boldsymbol{\mu}_{\boldsymbol{\theta}} - \overline{\mathbf{y}}_n||_2^2}{n \sigma^2_\omega}) \label{eqn:KL21}
\end{align}

\begin{figure}[htb]
\centering
\includegraphics[trim=20cm 0cm 20cm 0cm, scale=1]{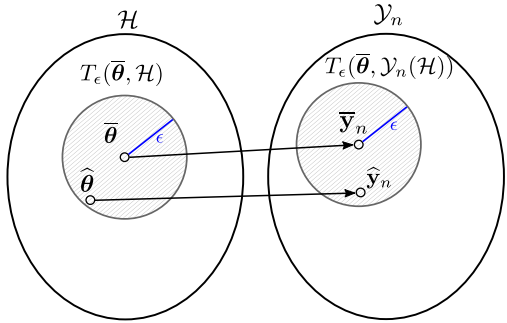}
\caption{$\epsilon$-Confidence Approximately Correct ($\epsilon$-CoAC) hypothesis set and $\epsilon$-CoAC target set $\overline{\boldsymbol{\theta}}$.}
\label{fig:MSE1}
\end{figure}

Therefore, $\epsilon$-CoAC typical set in the hypothesis class is equivalently represented by a set in the target vector space as follows:
\begin{align}
    T_{\epsilon}(\overline{\boldsymbol{\theta}},\mathcal{Y}_n(\mathcal{H})) = \left \{ \boldsymbol{\theta} \in \mathcal{H} : \frac{1}{2}(\frac{|| \boldsymbol{\mu}_{\boldsymbol{\theta}} - \overline{\mathbf{y}}_n||_2^2}{n \sigma^2_\omega}) < \epsilon \right \} \label{eqn:KL22}
\end{align}
As shown in (\ref{eqn:firstyhat}) $\widehat{\mathbf{y}}_n$ is equivalent to $\boldsymbol{\mu}_{\widehat{\theta}}$.
An estimator $\widehat{\boldsymbol{\theta}}$ is learnable with $\epsilon$ Confidence if and only if it belongs to the $\epsilon$-CoAC
set of $\overline{\boldsymbol{\theta}}$ which due to the structure of the KL distance in (\ref{eqn:KL21}) and (\ref{eqn:KL22}) is equivalent to the output estimate belonging to the $\epsilon$-CoAC target set of $\overline{\boldsymbol{\theta}}$ (as shown in Figure \ref{fig:MSE1}) :
\begin{align}
    \left [ \widehat{\boldsymbol{\theta}} \in T_{\epsilon}(\overline{\boldsymbol{\theta}}, \mathcal{H}) \right ] \;\;\; \boldsymbol{\equiv} \;\;\; \left[\widehat{\mathbf{y}}_n \in T_{\epsilon}(\overline{\boldsymbol{\theta}},\mathcal{Y}_n(\mathcal{H})) \right]
\end{align}
The equivalent notion of $\epsilon$-CoAC confirms the importance of the {\bf Noise-free MSE (NMSE)} Empirical Risk not only in the $\ell_2$ norm sense, but also as a measure of learnability:
\begin{align}
  r^{NMS} (\boldsymbol{\theta})= \frac{1}{n} \sum_{i=1}^{n} ({\mu}_{\boldsymbol{\theta}}(i) -\bar y(i))^2 \label{eqn:nlerSample}
 \end{align}
The NMSE that is calculated for the estimated parameter $\hat \theta$ is denoted as the \textbf{desired NSME (dNMSE)}, considering $\boldsymbol{\mu}_{\boldsymbol{\widehat{\theta}}}=\widehat{\mathbf{y}}$ in (\ref{eqnn:muAndy}), is calculated as: 
\begin{align}
     r^{N}(\widehat{\boldsymbol{\theta}})=r^{NMS} (\widehat{\boldsymbol{\theta}})= \frac{1}{n} \sum_{i=1}^{n} (\widehat{y}(i) -\overline{y}(i))^2 \label{eqn:newlerSample}
 \end{align} 
Note that the lengthy superscript $NMS$ which is comparable to the superscript $MS$ in (\ref{eqn:lerSample}) from now on is replaced by $N$. 

The estimate $\widehat{\boldsymbol{\theta}}$ is {\bf $\epsilon$-CoAC learnable} if:
\begin{align}
    \left [ \widehat{\boldsymbol{\theta}}  \in T_{\epsilon}(\overline{\boldsymbol{\theta}}, \mathcal{H}) \right ] \;\; &\boldsymbol{\equiv} \;\; \left [ r^{2N}(\widehat{\boldsymbol{\theta}})  < \epsilon \right ] \label{eqn:eplear1}\\
    r^{2N}(\widehat{\boldsymbol{\theta}}) &= \frac{1}{2 \sigma^2_{\omega}}r^N(\widehat{\boldsymbol{\theta}}) \label{eqn:CoacLearnable}
\end{align}
where $r^{2N}$ denotes the \textbf{desired Normalized Noise-free Mean Square Error (d2NMSE)}.

\section {$\epsilon$-CoAC Learnability as a function of data length and Hypothesis Class Complexity}
\label{sec:CoAC}
This section is dedicated to elaborating the $\epsilon$-CoAC learnability for regression models as a function of both data length and hypothesis class complexity. To compare hypothesis classes with different complexity, we define $\mathcal{H}_m$ as parametric model classes such that only the first $m$ elements of its parameter $\boldsymbol{\theta}$ are non-zero. 
\begin{align}
     \mathcal{H}_m \rightarrow \boldsymbol{\mu}_{m} &= \mathbf{A}_m\boldsymbol{\theta} \label{eqn:HMu}
\end{align}
where $\mathbf{A}_m$ is a full rank $m\times n$ matrix generated by the input vector. Note that the hypothesis class can have any parametric structures. For example, it can represent a set of parameters with only the third and seventh elements of $\boldsymbol{\theta}$ being non-zero. However, here for simplicity and without loss of generality, we analyze and study the behavior of estimators in the defined $\mathcal{H}_m$s. The value of $m$ can be any given value ranging between 1 and $n$, the length of the observed data. In a hypothesis class $\mathcal{H}_m$ with $m$ parameters and with the target vector of length $n$, the least square estimate of the true parameter in (\ref{eqn:OLS22}) is denoted as follows. 
\begin{align}
    \widehat{\boldsymbol{\theta}}_{m,n} &= \underset{\boldsymbol{\theta} \in \mathcal{H}_m}{\arg \min} \; r^{MS}(\boldsymbol{\theta}) =(\mathbf{A}_m^T\mathbf{A}_m)^{-1}\mathbf{A}_m^T\mathbf{y}_n \label{eqn:MSELS}
\end{align}
This element of the hypothesis class imposes the following probability distribution on the target set: 
\begin{gather}
    \widehat{\boldsymbol{\theta}}_{m,n}:\mathcal{D}_{\mathbf{y}}(\widehat{\boldsymbol{\theta}}) \sim \mathcal{N}(\boldsymbol{\mu}_{\widehat{\boldsymbol{\theta}}_{m,n}}, \sigma^2_\omega) \label{eqn:klybartheta2}\\
    \boldsymbol{\mu}_{\widehat{\boldsymbol{\theta}}_{m,n}}=\widehat{\mathbf{y}}_{m,n} =\mathbf{A}_m \widehat{\boldsymbol{\theta}}_{m,n} \label{eqn:muthetare}
\end{gather}
\subsection{Role of the MS Empirical Risk in the Parameter Estimation Process}
Here we introduce an important hypothesis corresponding directly to the observed noisy target vector $\mathbf{y}$. If we solve for the parameter estimate in $\mathcal{H}_n$, which is the hypothesis class with the highest possible order, that is, the length of the observed data, the MSE risk in (\ref{eqn:lerSample}) is minimized as follows  
\begin{align}
    \widehat{\boldsymbol{\theta}}_{n,n} = \underset{\boldsymbol{\theta} \in \mathcal{H}_n}{\arg \min} \; r^{MS}(\boldsymbol{\theta}) =(\mathbf{A}_n^T\mathbf{A}_n)^{-1}\mathbf{A}_n^T\mathbf{y}_n
    \label{eqn:OLS449}
\end{align}
As $\mathbf{A}_n$ is an $n\times n$ matrix and is full rank, the associated mean of this estimator is the same as $\mathbf{y}$ itself, i.e.,
\begin{align}
\boldsymbol{\mu}_{\widehat{\boldsymbol{\theta}}_{n,n}}  =\mathbf{A}_n \widehat{\boldsymbol{\theta}}_{n,n} = \widehat{\mathbf{y}}_{n,n} =\mathbf{y}_n
\end{align}
Therefore, the distribution generated by this estimator is:
 \begin{align}
     \widehat{\boldsymbol{\theta}}_{n,n}: \mathcal{D}_{\mathbf{y}}(\boldsymbol{\theta}_n) \sim \mathcal{N}(\mathbf{y}_n, \sigma^2_\omega)\label{eqn:ThetaNDist}
 \end{align}
This distribution is introduced for the following section to explain the role of the $\epsilon$- CoAC of this hypothesis in the complexity analysis. 
Therefore, the KL distance of this hypothesis and other elements of the hypothesis class $\mathcal{H}_m$ is
\begin{multline}
    D_{KL} \left ( \mathcal{D}_{\mathbf{y}}(\widehat{\boldsymbol{\theta}}_{n,n}) \parallel \mathcal{D}_{\mathbf{y}}(\boldsymbol{\theta}) \right ) = \\ \frac{1}{2}( \frac{||\mathbf{A}_{m}\boldsymbol{\theta} - {\mathbf{y}}_{n} ||_2^2}{n \sigma^2_\omega}) \label{eqn:KLNewNoisy} =\frac{1}{2 \sigma^2_\omega} r^{MS}({\boldsymbol{\theta}})
\end{multline}
Therefore, minimizing this KL distance equivalently results in minimizing the MSE (\ref{eqn:MSELS}):
\begin{align}
    \widehat{\boldsymbol{\theta}}_{m,n} 
    &= \underset{\boldsymbol{\theta} \in \mathcal{H}_m}{\arg \min} D_{KL} \left ( \mathcal{D}_{\mathbf{y}}(\widehat{\boldsymbol{\theta}}_{n,n}) \parallel \mathcal{D}_{\mathbf{y}}(\boldsymbol{\theta}) \right )
    \label{eqn:OLS33}
\end{align}
Using (\ref{eqn:klybartheta2}) and (\ref{eqn:ThetaNDist}) this minimum value is 
\begin{multline}
    D_{KL} \left ( \mathcal{D}_{\mathbf{y}}(\widehat{\boldsymbol{\theta}}_{n,n}) \parallel \mathcal{D}_{\mathbf{y}}(\widehat{\boldsymbol{\theta}}_{m,n}) \right ) = \\ \frac{1}{2}( \frac{||\widehat{\mathbf{y}}_{m,n} - {\mathbf{y}}_{n} ||_2^2}{n \sigma^2_\omega}) =\frac{1}{2 \sigma^2_\omega} r^{MS}(\widehat{\boldsymbol{\theta}}_{m,n})\label{eqn:KLNewChanged}
\end{multline}
\subsection{Role of the Noise-free Mean Square Error in $\epsilon$-CoAC learnability}
So far we did not have any assumption on the complexity of the 
true parameter. Assume that this value is $m^*$ which can be finite or even infinite in real applications. The value of $m^*$ in the true parameter $\overline{\boldsymbol{\theta}}_{m^*}$ is also assumed to be unknown in the considered learning process.    
The $\epsilon$-CoAC typical set in (\ref{eqn:COACOriginal}) for the hypothesis class of order $m$, $\mathcal{H}_m$, with a target set length of $n$ is then as follows:
\begin{multline}
    T_{\epsilon}(\overline{\boldsymbol{\theta}}_{m^*},\mathcal{Y}_n(\mathcal{H}_m)) = \\  \left \{ \boldsymbol{\theta} \in \mathcal{H}_m : D_{KL} \left ( \mathcal{D}_{\mathbf{y}}(\overline{\boldsymbol{\theta}}_{m^*}) \parallel \mathcal{D}_{\mathbf{y}} (\boldsymbol{\theta}) \right ) < \epsilon \right \} \label{eqn:TypicalFirst}
\end{multline}
The Kullback-Leibler distance between this true parameter
 and the estimated parameter in hypothesis class $\mathcal{H}_m$ in (\ref{eqn:OLS33}), using (\ref{eqn:klybartheta2}) and (\ref{eqn:klybartheta}) is
\begin{multline}
    D_{KL} \left ( \mathcal{D}_{\mathbf{y}}({\overline{\boldsymbol{\theta}}_{m^*}}) \parallel \mathcal{D}_{\mathbf{y}}(\widehat{\boldsymbol{\theta}}_{m,n}) \right ) = \\ \frac{1}{2}( \frac{||\widehat{\mathbf{y}}_{m,n} - {\overline{\mathbf{y}}_n} ||_2^2}{n \sigma^2_\omega}) = r^{2N}(\widehat{\boldsymbol{\theta}}_{m,n}) \label{eqn:klmnmn22}
\end{multline}

where $r^{2N}$ is defined in (\ref{eqn:CoacLearnable}).
Therefore, the estimate is $\epsilon$-CoAC learnable if this value is less than $\epsilon$. 

\begin{figure}[htb]
\centering
\includegraphics[trim=20cm 0cm 20cm 0cm, scale=1]{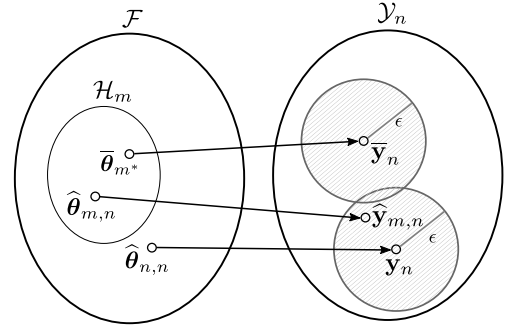}
\caption{Important three hypotheses in the learning process (tetas) and $\epsilon$-CoAC learnability}
\label{fig:MSE2}
\end{figure}

\begin{figure}[htb]
\centering
\includegraphics[trim=20cm 0cm 20cm 0cm, scale=1]{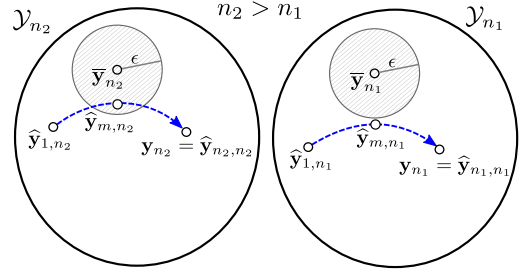}
\caption{$\epsilon$-CoAC learnability as the order of the hypothesis class increases from 1 ($\widehat{\mathbf{y}}_{1,n}$) to the highest possible order $n$ ($\widehat{\mathbf{y}}_{n,n}$) for two possible lengths of observations ($n_1$ and $n_2$)}
\label{fig:MSE3}
\end{figure}

Figure \ref{fig:MSE2} shows the three important hypotheses in this analysis, $\overline{\boldsymbol{\theta}}$, $\widehat{\boldsymbol{\theta}}_{n,n}$ and $\widehat{\boldsymbol{\theta}}_{m,n}$, as well as the important corresponding mean of the distribution associated with these three parameters $\overline{\mathbf{y}}_n$, ${\mathbf{y}}_n$, $\widehat{\mathbf{y}}_{m, n}$ and the corresponding $T_\epsilon$ with center $\overline{\mathbf{y}}$. The figure shows an example in which  $m^* \leq m$, i.e., the true parameter belongs to the hypothesis class $\mathcal{H}_m$, yet the estimator is not acceptable as it is outside of the desired Typical set. The figure also shows that if, by mistake, the Typical set is centered around the observed data, the estimate is in $\epsilon$-CoAC distance of the observed data. Figure \ref{fig:MSE3} illustrates the validation process as the order of $\mathcal{H}_m$ increases for two cases with two different data length $n_1$ and $n_2$, where $n_2>n_1$. As the figure shows, the center of the estimated distributions that corresponds to the estimated target vectors in the form of $\mathbf{y}_{m,n}$ converges to the observed data $\mathbf{y}_n = \widehat{\mathbf{y}}_{n,n}$, as the order increases. However, in the case of $n_1$ none of the estimates is $\epsilon$-CoAC learnable. As the data length increases to $n_2$, however, there is a set of parameters of order $m$ that is $\epsilon$-CoAC learnable. Note that the convergence of the estimators to the observed data itself is a known Maximum Likelihood (ML) issue in complexity analysis \cite{domingos2000bayesian, hastie2009elements, myung2006model}. Here the LS estimate, $\widehat{\boldsymbol{\theta}}$, is also the ML estimate and if the goal is to minimize the MSE, the method always chooses the estimator with the highest possible order. 
However, comparison of $\epsilon$-CoAC learnability can produce different results. For calculation of this learnability, the NMS is needed. In the following sections, we provide a method for estimating the unavailable NMS Empirical risk using the available MS Empirical risk of the estimator. We first discuss the case that the unavailable true order $m^*$ is known to be a member of the hypothesis class, $m > m^*$, and address the $\epsilon$-CoAC learnability as a function of the data length. Next, this assumption is relaxed, and the general $\epsilon$-CoAC learnability is discussed as a function of data length and hypothesis class order. Note that from now on the two errors,  the available minimum MS Empirical risk, $r^{MS}(\widehat{\boldsymbol{\theta}}_{m,n})$ in (\ref{eqn:rm}) and the unavailable $r^{N}_{m,n}$ in (\ref{eqn:TypicalFirst}) will be used frequently. For simplicity we will shorten their notations as follows
\begin{align}
    r^{MS}_{m,n} = r^{MS}(\widehat{\boldsymbol{\theta}}_{m,n})
\end{align}
\begin{align}
    r^{N}_{m,n} = r^N(\widehat{\boldsymbol{\theta}}_{m,n})
\end{align}

\section{$\epsilon$-CoAC Regression Learnability when $m^* \leq m$ (Model Belongs to Hyothesis Class)}
\label{section:SaCo}
In this section, we assume that the unknown true parameter $\overline{\boldsymbol{\theta}}_{m^*}$ belongs to the hypothesis class with complexity order $\mathcal{H}_m$. This assumption is relaxed in the next section. The focus of this section is on $\epsilon$-CoAC learnability as a function of data length $n$. Using the available mMSE, we provide probabilistic bounds on the dNMSE for the learnability process.

Using (\ref{eqn:MSELS}) and (\ref{eqn:noisyoutput}), the least squares estimate of the parameter vector when $m^*\leq m$ is:
\begin{align}
    \widehat{\boldsymbol{\theta}}_{m, n} &=\left ( (\mathbf{A}_m^T\mathbf{A}_m)^{-1}\mathbf{A}_m^T \right )
    \left (\overline{\mathbf{y}}_n + \boldsymbol{\omega} \right ) = \overline{\boldsymbol{\theta}}_{m} + (\mathbf{A}_m^T\mathbf{A}_m)^{-1}\mathbf{A}_m^T \boldsymbol{\omega} \label{eqn:tetahatinnn}
\end{align}
This will lead to the following dNMSE in (\ref{eqn:klmnmn22})
as $\overline{\mathbf{y}}_m = \mathbf{A}_m \overline{\boldsymbol{\theta}}_m$
\begin{align}
    r^{N}_{m, n} &= \frac{1}{n} \left \| \mathbf{A}_m \widehat{\boldsymbol{\theta}}_{m, n} - \mathbf{A}_m \overline{\boldsymbol{\theta}}_{m} \right \|^2_2 \label{eqn:rmnsefirstline} =\frac{1}{n} \left \|\mathbf{A}_m (\mathbf{A}_m^T\mathbf{A}_m)^{-1}\mathbf{A}_m^T \boldsymbol{\omega} \right \|^2_2 \\
    &=\frac{1}{n} \left \|\mathbf{H}_m \boldsymbol{\omega} \right \|^2_2 = \frac{1}{n} (\mathbf{H}_m \boldsymbol{\omega}_n)^T (\mathbf{H}_m \boldsymbol{\omega}) \label{eqn:Hmforn-1} \\
    &= \frac{1}{n} \boldsymbol{\omega}^T \mathbf{H}_m \boldsymbol{\omega} \label{eqn:Hmforn}
\end{align}
where from (\ref{eqn:Hmforn-1}) to (\ref{eqn:Hmforn}) the fact that $\mathbf{H}_m$, the hat matrix:  
\begin{align}
    \mathbf{H}_m = \mathbf{A}_m (\mathbf{A}_m^T\mathbf{A}_m)^{-1}\mathbf{A}_m^T \label{eqn:projection}
\end{align}
is a projection matrix of rank $m$ is used. Due to the structure of $r^{N}_{m, n}$ in (\ref{eqn:Hmforn}), it is a sample of a chi-squared random variable denoted by $R^{N}_{m, n}$ with the following mean and variance \cite{patel1996handbook}:
\begin{align}
    \mathbb{E}(R^{N}_{m, n}) &= \frac{1}{n} \mathbb{E}\left ( \boldsymbol{\omega}^T \mathbf{H}_m \boldsymbol{\omega} \right )= \frac{m}{n} \sigma^2_\omega \label{eqn:EVTm1}\\
    \mathrm{var}(R^{N}_{m, n}) &= \frac{1}{n^2} \mathrm{var}\left ( \boldsymbol{\omega}^T \mathbf{H}_m \boldsymbol{\omega} \right )= \frac{2m}{n^2}(\sigma_\omega^2)^2 \label{eqn:EVTm}
\end{align}
While the dNMSE itself is unknown, the Chebyshev inequality can provide probabilistic bounds on this value.
Using Chebyshev's inequality for $R^{N}_{m, n}$, the value of the dNMSE in (\ref{eqn:rmnsefirstline}) is bounded around the mean with probability $P_\beta = 1 - \frac{1}{\beta^2}$ as follows:
\begin{align}
    P\left ( \left | \mathbb{E}(R^{N}_{m, n}) - r^{N}_{m, n} \right | \leq \beta \sqrt{\mathrm{var}(R^{N}_{m, n})}\right ) \geq P_\beta \label{eqn:Pz}
\end{align}
and, therefore, we have
\begin{align}
    \mathbb{E}(R^{N}_{m, n}) - \beta \sqrt{\mathrm{var}(R^{N}_{m, n})} \leq r^{N}_{m, n} \leq  \mathbb{E}(R^{N}_{m, n}) + \beta \sqrt{\mathrm{var}(R^{N}_{m, n})} \label{eqn:RMNB}
\end{align}
$P_\beta$ and, consequently, $\beta$ determine the confidence of the probabilistic bounds in (\ref{eqn:RMNB}). 
Using values of the mean and variance from (\ref{eqn:EVTm1}) and (\ref{eqn:EVTm}), with confidence probability $P_\beta$, the unavailable dNMSE error is bounded as follows:
\begin{align}
    \underline{r^{N}_{m, n}} \leq r^{N}_{m, n} \leq \overline{r^{N}_{m, n}} \label{eqn:boundsfornep}
\end{align}
where 
\begin{align}
    \underline{r^{N}_{m, n}} &=  \frac{m-\beta \sqrt{2m}}{n} \sigma^2_\omega \\
    \overline{r^{N}_{m, n}} &= \frac{m+\beta \sqrt{2m}}{n} \sigma^2_\omega \label{eqn:upperrmnnew}
\end{align}

\begin{figure}[htb]
\centering
\includegraphics[trim=20cm 0cm 20cm 0cm, scale=1.3]{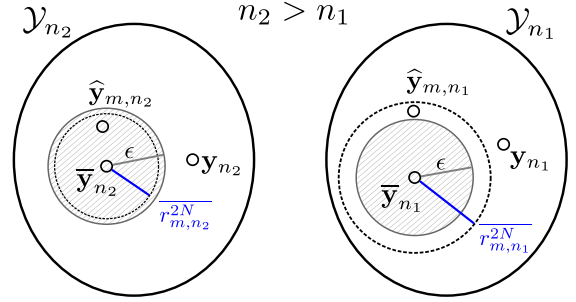}
\caption{$\epsilon$-CoAC learnability using the calculated upper bound on d2NMSE, for two data lengths $n_1$, $n_2$, $n_2 > n_1$. Here, $\widehat{\mathbf{y}}_{m,n_1}$ is not $\epsilon$-CoAC learnable while $\widehat{\mathbf{y}}_{m,n_2}$ is $\epsilon$-CoAC learnable.}
\label{fig:nComplexity}
\end{figure}

Using the upperbound, the $\epsilon$-CoAC learnability in (\ref{eqn:eplear1}) and (\ref{eqn:CoacLearnable}), with confidence probability of $P_\beta$, the $\widehat{\boldsymbol{\theta}}_{m,n}$ is $\epsilon$-CoAC learnable if:
\begin{align}
   \overline{r^{2N}_{m, n}} \leq \epsilon \rightarrow
    n \geq \frac{m + \beta \sqrt{2m}}{2 \epsilon} \label{eqn:generalNLearn}
\end{align}
Figure \ref{fig:nComplexity} illustrates the role of $\overline{r^{2N}_{m,n}}$, the calculated probabilistic worst case bound of d2NMSE in the $\epsilon$-CoAC framework. The figure shows the learnability for two values of the length of the training set, $n_1$ and $n_2$, where $n_1$ is smaller than $n_2$. As the figure shows, in this scenario the observed target vectors $\mathbf{y}_{n_1}$ and $\mathbf{y}_{n_2}$  are not inside the desired $\epsilon$-CoAC Typical set, $T_{\epsilon}(\overline{\boldsymbol{\theta}}_{m^*},\mathcal{Y}_n(\mathcal{H}_m))$, which are centered by $\overline{\mathbf{y}}_{n_1}$ and $\overline{\mathbf{y}}_{n_2}$ respectively. The distance between the noise-free target vectors $\overline{\mathbf{y}}_{n_1}$, $\overline{\mathbf{y}}_{n_2}$, and the estimated target vector $\widehat{\mathbf{y}}_{m, n_1}$, $\widehat{\mathbf{y}}_{m, n_2}$, determine the desired unavailable dNMSE in the $\epsilon$-CoAC framework.
 On the other hand, the circle with $\overline{r^{2N}_{m, n}}$ radius represents the probabilistic worst case bound of ${r^{2N}_{m, n}}$ that can represent the minimum acceptable value of $\epsilon$ with confidence $P_\beta$ as:
\begin{align}
    \overline{r^{2N}_{m, n_2}} < \epsilon < \overline{r^{2N}_{m, n_1}}
\end{align}
It can be said that with confidence probability $P_\beta$ the estimator for sample length $n_2$ is $\epsilon$-CoAC learnable, while with the same confidence probability the estimator from sample length $n_1$ is not  $\epsilon$-CoAC learnable.
\subsection{Noise Variance Estimation When $\overline{\boldsymbol{\theta}}_{m^*} \in \mathcal{H}_m$}
Usually in practical applications no prior information about the noise variance, and $\sigma^2_\omega$ is available. 
In this case the structure of the available $r^{MS}$ enable estimating the noise variance. Using (\ref{eqn:OLS22}) and (\ref{eqn:noisyoutput}), $r_{m, n}^{MS}$ in (\ref{eqn:rm}) can be rewritten as follows:
\begin{align}
    r^{MS}_{m, n} &= \frac{1}{n} \left \| \mathbf{A}_m \widehat{\boldsymbol{\theta}}_{m, n} - (\mathbf{A}_m \overline{\boldsymbol{\theta}}_{m, n} + \boldsymbol{\omega}) \right \|^2_2  \label{eqn:unlike}\\
    &= \frac{1}{n} \left \|(\mathbf{H}_m - \mathbf{I}_m) \boldsymbol{\omega} \right \|^2_2 =\frac{1}{n} \left \|\mathbf{G}_m \boldsymbol{\omega} \right \|^2_2   \label{eqn:rmsss2}\\
    &= \frac{1}{n}\boldsymbol{\omega}^T \mathbf{G}_m \boldsymbol{\omega} \label{eqn:rmsss}
\end{align}
where from (\ref{eqn:rmsss2}) to (\ref{eqn:rmsss}) we use the fact that matrix $\mathbf{G}_m$ 
\begin{align}
    \mathbf{G}_m = \mathbf{H}_m - \mathbf{I}_m \label{eqn:GMProjec}
\end{align}
is a projection matrix of rank $n-m$. Due to the structure of $r^{MS}_{m, n}$ in (\ref{eqn:rmsss}), it is a sample of a chi-squared random variable denoted by $R^{MS}_{m, n}$ with the following mean and variance \cite{patel1996handbook}:
\begin{align}
    \mathbb{E}(R^{MS}_{m, n}) &= \frac{1}{n} \mathbb{E}\left ( \boldsymbol{\omega}^T \mathbf{G}_m \boldsymbol{\omega}\right ) = (1 - \frac{m}{n})\sigma^2_\omega \label{eqn:EER_m}\\
    \mathrm{var}(R^{MS}_{m, n}) &=\frac{1}{n^2}\mathrm{var}\left ( \boldsymbol{\omega}^T \mathbf{G}_m \boldsymbol{\omega}\right ) =\frac{2}{n}(1 - \frac{m}{n})(\sigma^2_\omega)^2 \label{eqn:EVR_m}
\end{align}
Existing approaches for noise variance estimation equate the mean value in (\ref{eqn:EER_m}) with the available sample $r^{MS}_{m,n}$ and provide the following noise variance estimate\cite{guo2011estimation, montgomery2021introduction}: 
\begin{align}
    \widehat{\sigma}^2_{\omega} \approx \frac{n}{n-m}  r^{MS}_{m,n}
\end{align}
However this estimate ignores the behavior of the variance of $R^{MS}_{m,n}$ which itself is a function of $m$ and $n$. Therefore, as the value of $n$ changes, the reliability of this estimate varies. To avoid such issue, we propose to estimate noise variance with a fixed value of validation probability as follows. Using the Chebyshev inequality for $R^{MS}_{m,n}$, we have
\begin{align}
    P\left ( \left | \mathbb{E}(R^{MS}_{m, n}) - r^{MS}_{m, n} \right | \leq \alpha \sqrt{\mathrm{var}(R^{MS}_{m, n})}\right ) \geq P_\alpha \label{eqn:Pn}
\end{align}
where $P_\alpha = 1 - \frac{1}{\alpha^2}$. The probability $P_\alpha$ and consequently $\alpha$ determine the validation probability for the bounds on $R^{MS}_{m,n}$. Please note that previously in (\ref{eqn:RMNB}) we used the Chebyshev inequality to find probabilistic bounds on the unavailable sample $r^{N}_{n,m}$ with a confidence probability $P_\beta$. Here the Chebyshev inequality is not used to find confidence bounds, but is used to validate values of $\sigma^2_\omega$ that satisfy the inequality.  Therefore, with the validation probability $P_\alpha$, the calculated $r^{MS}_{m,n}$ is bounded by a range of possible $\sigma_\omega^2$. In Appendix A we provide details of using the above Chebyshev inequality to validate the following range for $\sigma^2_\omega$:
\begin{align}
     \frac{n r^{MS}_{m, n}}{n -m+ \alpha \sqrt{2n - 2m}} \leq  \sigma^2_\omega \leq  \frac{n r^{MS}_{m, n}}{n -m- \alpha \sqrt{2n - 2m}} \label{eqn:rmenqnn}
\end{align}
Interestingly this procedure allows the method to provide an estimate for the noise-free mean square error ($r^N_{m,n}$).
Therefore, with confidence probability $P_\beta$ and validation probability $P_\alpha$, the dNMSE in (\ref{eqn:boundsfornep}) is bounded as follows:

\begin{align}
     \frac{(m - \beta \sqrt{2m}) (r^{MS}_{m, n})}{(n -m- \alpha \sqrt{2n - 2m})} \leq {r^N_{m,n}} \leq \frac{(m + \beta \sqrt{2m}) (r^{MS}_{m, n})}{(n -m+ \alpha \sqrt{2n - 2m})} \label{eqn:upperNNN}
\end{align}
These bounds are valuable in practical applications and in comparison of the performance of the learning algorithm as the data length changes. 
\section{$\epsilon$-CoAC Hypothesis Class Learnability}
\label{section:HyCa}
This section investigates learnability of a hypothesis class for a regression problem within the context of the $\epsilon$-CoAC framework. In the previous section it was known that the true parameter $\overline{\boldsymbol{\theta}}_{m^*}$  is a member of the considered hypothesis class $\mathcal{H}_m$. Here, this assumption is relaxed. Therefore, not only the true complexity order of the regression problem denoted by $m^*$, is unknown; but also it is not known whether $m^*\leq m$. 
The goal is to compare the parameter estimates in model classes of $m$, $\widehat{\boldsymbol{\theta}}_{m,n}$, for a range of $m$  $m \in {1, \dots, M}$. The value of $M$ can be any number less than or equal to the data length $M \leq n$ and is usually provided in practical applications. Here, the true parameter and the noise-free target vector in (\ref{eqn:ybar}) are
 \begin{align}
    \overline{\boldsymbol{\theta}}_{M} =  [ \underset{\overline{\theta}_{m^*}}{\underbrace{\bar \theta(1), \dots, \bar \theta(m^*),\;}} \underset{M - {m^*}}{\underbrace{0, \dots, 0}}]^T
    \label{eqn:noislessParam}
\end{align}
\begin{align}
    \overline{\mathbf{y}}_n = \mathbf{A}_M \overline{\boldsymbol{\theta}}_{M} = \mathbf{A}_{m^*} \overline{\boldsymbol{\theta}}_{m^*} \label{eqn:ynbarinHm}
\end{align}
The ordinary LS estimate of the true parameter in Hypothesis class with order $m$, $\widehat{\boldsymbol{\theta}}_{m,n}$ in (\ref{eqn:MSELS}), is:
\begin{multline}
    \widehat {\boldsymbol{\theta}}_{m, n} = \begin{bmatrix}
(\mathbf{A}^T_m \mathbf{A}_m)^{-1}\mathbf{A}^T_m {\mathbf{y}}_n\\ 
\mathbf{0}_{(M-m)\times 1}
\end{bmatrix} = \\
\begin{bmatrix}
\overline{\boldsymbol{\theta}}_m + (\mathbf{A}^T_m \mathbf{A}_m)^{-1}\mathbf{A}^T_m(\mathbf{B}_m \boldsymbol{\Delta}_m + \omega_n )\\ 
\mathbf{0}_{(M-m)\times 1}
\end{bmatrix} \label{eqn:tetahat_matrix}
\end{multline}
where $\mathbf{A}_m$ is the first $m$ columns of $\mathbf{A}_M$ and $\mathbf{B}_m$ is the columns $m+1$ up to $m^*$ of $\mathbf{A}_M$ and ${\boldsymbol{\Delta}}_{m}$ are the $(m^*-m)$ parameters that are not included in $\overline{\boldsymbol{\theta}}_{m}$. As a result, if $m<m^*$, the unmodeled parameters $\Delta_m$ have the following structure.
\begin{align}
    \overline{\boldsymbol{\theta}}_{M} = [\underset{\overline{\theta}_{m}}{\underbrace{\bar \theta(1), \dots, \bar \theta(m),\;}}\underset{\boldsymbol{\Delta}_{m}}{\underbrace{\bar \theta(m+1), \dots, \bar \theta(m^*),\; 0, \dots, 0}}]^T
    \label{eqn:deltaV1}
\end{align}
Note that the unmodeled $\boldsymbol{\Delta}_m$ is zero for when $m \geq m^*$, ${\boldsymbol{\Delta}}_{m} = \mathbf{0}_{(M-m) \times 1}$. Using (\ref{eqn:muthetare}), the estimated target vector corresponding to the estimated parameter vector in (\ref{eqn:tetahat_matrix}) is
\begin{align}
    \widehat{\mathbf{y}}_{m, n} = \mathbf{A}_M \widehat{\boldsymbol{\theta}}_{m, n} = \begin{bmatrix}
\mathbf{A}_m & \mathbf{B}_m 
\end{bmatrix}\widehat{\boldsymbol{\theta}}_{m, n} \label{eqn:AmBm}
\end{align}
While this value can be used to calculate mMSE, the dNMSE for $\epsilon$-CoAC learnability analysis is not available. Here, we derive probabilistic bounds on $r^N_{m, n}$ using the available $r^{MS}_{m, n}$. This calculation can be used not only for the $\epsilon$-CoAC learnability but also to determine the optimal hypothesis class for the given training data set. 

The unavailable dNMSE,  $r^{N}_{m,n}$, in (\ref{eqn:klmnmn22}),
using (\ref{eqn:ynbarinHm}), (\ref{eqn:tetahat_matrix}) and (\ref{eqn:AmBm}) is
\begin{align}
    r^N_{m, n} &= \frac{1}{n} \| \mathbf{A}_M ({\widehat{\boldsymbol{\theta}}}_{m, n} -   \overline{\boldsymbol{\theta}}_M)  \|_2^2 = \frac{1}{n} \| \mathbf{G}_m \mathbf{B}_m \Delta_m + \mathbf{H}_m \boldsymbol{\omega} \|_2^2  \label{eqn:1termsMon} \\
    &= \frac{1}{n} \| \mathbf{G}_m \mathbf{B}_m \Delta_m \|_2^2 + \| \mathbf{H}_m \boldsymbol{\omega} \|_2^2 \label{eqn:2termsMon}
\end{align}
where projection matrices $\mathbf{H}_m$ and $\mathbf{G}_m$ with rank $m$ and $n-m$ respectively, are defined in (\ref{eqn:projection}) and (\ref{eqn:GMProjec}). Note that going from (\ref{eqn:1termsMon}) to (\ref{eqn:2termsMon}) uses the fact that $\mathbf{H}_m$ and $\mathbf{G}_m$ are orthogonal:
\begin{align}
    \mathbf{G}_m \mathbf{H}_m = (\mathbf{H}_m - \mathbf{I}_m)\mathbf{H}_m = \mathbf{H}_m^2 - \mathbf{H}_m = 0
\end{align}
It is important to mention that as shown in Figure 
(\ref{fig:MSE3}), this error, for a fixed value of the data length $n$, is a convex function of $m$ with an absolute global minimum. This is due to the behavior of the two terms in (\ref{eqn:2termsMon}). While the first term, $ \| \mathbf{G}_m \mathbf{B}_m \Delta_m \|_2^2$, in (\ref{eqn:2termsMon}) is a monotonically decreasing function of $m$, the second term, $ \| \mathbf{H}_m \boldsymbol{\omega} \|_2^2$, is a monotonically increasing function of $m$. Therefore, there is always a value of $m$ that minimizes the unavailable $r^N_{m,n}$ and results in the best possible achievable accuracy in terms of NMSE.

Taking into account (\ref{eqn:noisyoutput}) and (\ref{eqn:klybartheta2}),  $r^N_{m, n}$ is a sample of a chi-squared random variable $R^N_{m, n}$ with the following mean and variance \cite{beheshti2009noisy}.
\begin{align}
    {\rm E}(R^N_{m, n})&=\frac{m}{n}\sigma_{\omega}^2+\frac{1}{n}||\mathbf{G}_{m}\mathbf{B}_{m}\boldsymbol{\Delta}_{m}||_2^2 \label{eqn:changedbounds}\\
{\rm var}(R^N_{m, n})&=\frac{2m}{n^2}(\sigma_{\omega}^2)^2.
\end{align}

On the other hand, the available $r^{MS}_{m, n}$, using (\ref{eqn:noisyoutput}), (\ref{eqn:tetahat_matrix}) and (\ref{eqn:AmBm}) is
\begin{align}
    r^{MS}_{m, n} = \frac{1}{n} \| \mathbf{A}_M ({\widehat{\boldsymbol{\theta}}}_{m, n} -   \overline{\boldsymbol{\theta}}_M) - \boldsymbol{\omega}  \|_2^2 = \frac{1}{n} \| \mathbf{G}_m \mathbf{B}_m \Delta_m + \mathbf{G}_m \boldsymbol{\omega} \|_2^2
\end{align}
Using (\ref{eqn:noisyoutput}) and (\ref{eqn:OLS33}), $r^{MS}_{m, n}$ is a sample of a Chi-Square random variable $R^{MS}_m$ with the following mean and variance \cite{beheshti2009noisy}  follows:
\begin{align}
    \mathrm{E}(R^{MS}_{m, n})&=(1-\frac{m}{n})\sigma_\omega^2+\frac{1}{n}||\mathbf{G}_{m}\mathbf{B}_{m}\boldsymbol{\Delta}_{m}||_2^2\\
{\rm var}(R^{MS}_{m, n})&=\frac{2}{n}(1-\frac{m}{n})(\sigma_\omega^2)^2+
\frac{4\sigma_\omega^2}{n^2}||\mathbf{G}_{m}\mathbf{B}_{m}\boldsymbol{\Delta}_{m}||^2_2
\end{align}

\subsection{Probabilistic Structure of dNMSE and mMSE, Probabilistic Bounds on the Desired $r^N_{m,n}$}

Similarly to the approach proposed in Section \ref{sec:CoAC}, an upper bound for the unavailable $r^N_{m,n}$ can be provided with confidence probability $P_\beta$. While the bounds in (\ref{eqn:RMNB}) are only  functions of the noise variance $\sigma^2_\omega$, here the bounds are functions of both $\sigma^2_\omega$
and the unmodeled parameters effect $\boldsymbol{\Delta}_m$ which is also unknown. This effect of the unknown unmodeled parameters can also be probabilistically validated by considering the structure of mean and variance of the random variable $R^{MS}_{m,n}$. The value of $\boldsymbol{\Delta}_m$ is first probabilistically validated using Chebyshev's inequality on the available sample, $r^{MS}_{m,n}$ for each $m$. The resulting bounds on this value are then used in the first Chebyshev's confidence probabilistic bounds on the unavailable $r^N_{m,n}$. Details of this method of probabilistic calculation of bounds are similar to the probabilistic validation in Linear Time Invariant model estimation in \cite{beheshti2009noisy, shamsi2022relative}. 
The estimate of the unknown noise variance in this scenario can also be included in this procedure \cite{beheshti2009noisy, shamsi2022relative}. Consequently, the probabilistic bounds of $R^N_{m, n}$ with confidence probability $P_\beta$ and validation probability $P_\alpha$ are:
\begin{multline}
    \overline{r^N_{m, n}} = r^{MS}_{m, n} + \kappa (\alpha, n, m , \sigma_\omega^2) + \frac{2 \alpha^2 \sigma_\omega^2}{n} - \eta(m, n, \sigma_\omega^2) + \\ \frac{m}{n} \sigma^2_\omega + \beta \frac{\sqrt{2m} \sigma^2_\omega}{n} \label{eqn:finalUpper}
\end{multline}
\begin{multline}
    \underline{r^N_{m, n}} = r^{MS}_{m, n} - \kappa (\alpha, n, m , \sigma_\omega^2) + \frac{2 \alpha^2 \sigma_\omega^2}{n} - \eta(m, n, \sigma_\omega^2) + \\ \frac{m}{n} \sigma^2_\omega - \beta \frac{\sqrt{2m} \sigma^2_\omega}{n} \label{eqn:finalower}
\end{multline}
where
\begin{align}
    \kappa (\alpha, n, m , \sigma_\omega^2) &= 2 \alpha \frac{\sigma_\omega}{n} \sqrt{ \frac{\alpha ^2 \sigma^2_\omega}{n} + r^{MS}_{m, n} - \frac{\eta(m, n, \sigma_\omega^2)}{2} } \\
    \eta(m, n, \sigma_\omega^2) &= (1-\frac{m}{n})\sigma^2_\omega \label{eqn:upperHm}
\end{align}
\subsection{$\epsilon$-CoAC learnability, complexity selection, and data driven minimum $\epsilon$ in learnability}
Using the estimated upper bound, dNMSE from (\ref{eqn:finalUpper}), the d2NMSE  $\overline{r^{2N}_{m,n}}$ in (\ref{eqn:CoacLearnable}), can be written as:
\begin{multline}
    \overline{r^{2N}_{m,n}} =r^{nrMS}_{m,n} + \frac{2 \alpha}{n}\sqrt{\frac{\alpha^2}{n} + r^{nrMS}_{m,n} - \frac{1}{2}(1 - \frac{m}{n})} + \\ \frac{2 \alpha^2 + 2m + \beta \sqrt{2m} - n}{n} \label{eqn:r2nwithrmsn}
\end{multline}
where the normalized MSE denoted by nrMSE is defined as
\begin{align}
    r^{nrMS}_{m,n} = \frac{r^{MS}_{m,n}}{\sigma^2_\omega}
\end{align}
The estimated parameter in $\mathcal{H}_m$, $\widehat{\boldsymbol{\theta}}_{m,n}$ is $\epsilon$-CoAC  (\ref{eqn:klmnmn22}) learnable, with confidence probability of $P_\beta$, and validation probability of $P_\alpha$ if for the associated probabilistic worst-case upper bound  $\overline{r^{2N}_{m,n}}$ in (\ref{eqn:r2nwithrmsn}) we have:
\begin{align}
   \overline{r^{2N}_{m, n}} \leq \epsilon \label{eqn:coaconn222}
\end{align}

The probabilistic worse case upperbounds on the desired $r^{2N}_{m,n}$ can also be used for comparison of model classes of different complexity. The model class with the smallest upper bound is the best among them in the sense of $\epsilon$-CoAC learnability. For example, in comparison of model classes defined in this section with complexity ranging from 1 to $M$, the one that performs the best based on the available data in this sense is
\begin{align}
    \widehat{\boldsymbol{\theta}}_{\widehat{m}^*}(n) = \underset{1 \leq m \leq M} {\arg \min} (\overline{r^{2N}_{m, n}}) \label{eqn:FullOptimization}
\end{align}
Consequently, the associated minimum possible value of such an upper bound at $\widehat{m}^*$ is equivalently the minimum possible $\epsilon$ that can choose at least one member of the competing hypothesis classes.
\begin{align}
    \epsilon_{min}(n,m, P_\alpha, P_\beta, S) = \overline{r^{2N}_{\widehat{m}^*, n}} \label{eqn:minepsilon}
\end{align}

\subsection{Validation of $M$, the Higher Level Hyperparameter}
That upper bound of hypothesis class complexity order, denoted by $M$ in (\ref{eqn:noislessParam}), in general case can be a number as large as the training data length $n$. However, in practical applications, for two possible reasons, an upper bound $M$ is considered. In some cases, a large enough $M$ is available as prior information about the complexity order of the hypothesis class. On the other hand, in some practical applications, only due to complexity analysis, a prior $M$ is considered as a potential upper bound for $m^*$. In this scenario, the assumption in (\ref{eqn:noislessParam}), $m^* \in [1, M]$, may not be a valid assumption. Currently, no method of validation for this assumption is available. The behavior of the calculated  proposed $\overline{r^N_{m,n}}$ is however a solid measure for this validation. It is important to note that the calculation of this probabilistic bound is not a function of the considered $M$. Therefore, it is not affected by this value. In addition, $\overline{r^N_{m,n}}$  is a convex function of $m$. While $r^{MS}_{m,n}$, the first term of $\overline{r^N_{m,n}}$ in (\ref{eqn:finalUpper}), is a decreasing function of $m$, the sum of the other terms is a monotonically increasing function of $m$, and this guarantees the existence of a global minimum. Consequently, if minimization in (\ref{eqn:FullOptimization}) results in $\widehat{m}^*=M$ this indicates two possibilities. 
Either $M$ is the true value that minimizes the desired error, or there is an order higher than $M$ that further minimizes this criterion. Based on this observation, the $\epsilon$-CoAC learner has two options: Either $\overline{r^{2N}_{m,n}}<\epsilon$  which indicates that the desired learnability is achieved and, therefore, the learning procedure stops, or $\overline{r^{2N}_{m,n}}>\epsilon$ which indicates that the estimator is not $\epsilon$-CoAC learnable. In the latter case, the learner can increase the value of $M$ up to a value for which $\overline{r^{2N}_{m,n}}$  starts to increase, which means that the method has found the absolute best possible complexity with $ \epsilon_{\min}$ in (\ref{eqn:minepsilon}). In any case, if the goal of the learner is to achieve the minimum possible $\epsilon$ which is $\epsilon_{\min}$, the value of $M$ must be increased to where the global minimum is achieved.

\section{Simulation Results}
\begin{figure}[htb]
\centering
\includegraphics[trim=0cm 0cm 0cm 0cm, scale=0.38]{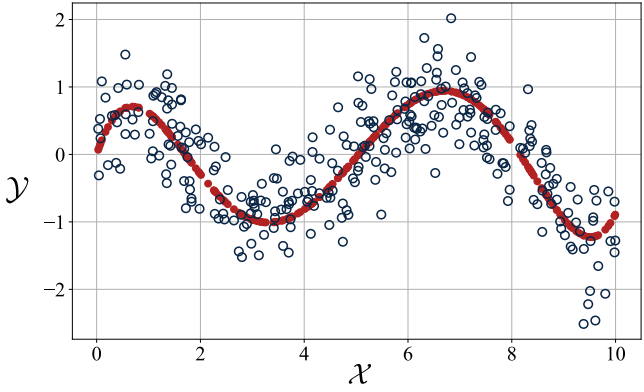}
\caption{Training data set and the respective true target vector.}
\label{fig:trainigdata}
\end{figure}

Here the $\epsilon$-CoAC learnability and the proposed optimal hypothesis class complexity selection methods are examined on a polynomial regression example. The desired $ f(\mathbf{x})$ is a polynomial of order 5 with the following parameters:
\begin{align}
    \overline{\boldsymbol{\theta}}_{m^*} = [2.3348, -2.3403, 0.6988, -0.0809, 0.0032]^T \label{eqn:simTheta}
\end{align}
The input vector $\mathbf{x}$ in (\ref{eqn:noisyoutput}) is generated by the sampling interval $(0, 10)$ uniformly. The kernel function of the regression is $\phi({x_i}) = [x_i, x_i^2, \dots, x_i^5]$. Figure \ref{fig:trainigdata} shows an example of the true target vector $\overline{\mathbf{y}}$ in (\ref{eqn:noisyoutput}) and the available noisy target vector $\mathbf{y}$ with length $n=200$ and noise variance $\sigma^2_\omega = 0.2$.

\subsection{$\epsilon$-CoAC Learnability as a Function of the Training Data Set Length when $\overline{\boldsymbol{\theta}}_{m^*} \in \mathcal{H}_m$}
\begin{figure}[h]
\centering
\includegraphics[trim=0cm 0cm 0cm 0cm, scale=0.41]{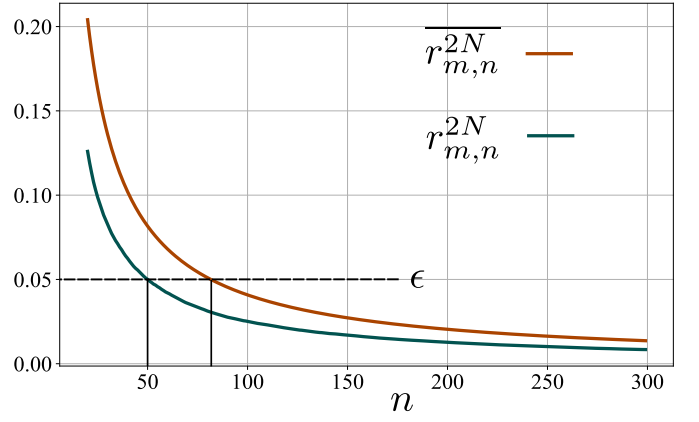}
\caption{$r^{2N}_{m, n}$ and estimated $\overline{r^{2N}_{m, n}}$ for known $m^*$ with validation probability $P_\beta=95.45\%$.}
\label{fig:rmnforn2}
\end{figure}

In the first experiment it is assumed that the true order is known to be 5 ($m^*=5$). For this scenario, Section \ref{section:SaCo} is followed to analyze $\epsilon$-CoAC complexity as a function of data length.  Figure \ref{fig:rmnforn2} shows the desired unavailable ${r^{2N}_{m, n}}$ and the estimated $\overline{r^{2N}_{m, n}}$ in (\ref{eqn:generalNLearn}). 

The results are for data lengths from 15 to 300 averaged over 1000 runs of the algorithm. The figure also shows the minimum value of $n$ for $\epsilon$-CoAC learnability when $\epsilon=0.05$. While the minimum $n$ based on the desired unavailable d2NMSE is $n=50$, the estimated probabilistic d2NMSE, $\overline{r^{2N}_{m,n}}$, with validation probability $P_\beta=95.45\%$, provides value of $n=81$.
Table \ref{tab:nsampleTable2} shows the required number of samples in $\epsilon$-CoAC learnability for different values of $\epsilon$ averaged over 1000 trials. The column shown under $n (\overline{r^{2N}_{m,n}})$ is the result of the estimated upper bound in (\ref{eqn:generalNLearn}). As the table shows, the estimated values of the data length with the proposed approach are valid estimates of the true number of samples using the unavailable d2NMSE. As expected, when the value of $\epsilon$ decreases, the required data length also increases.

\begin{figure}[h]
\centering
\includegraphics[trim=0cm 0cm 0cm 0cm, scale=0.41]{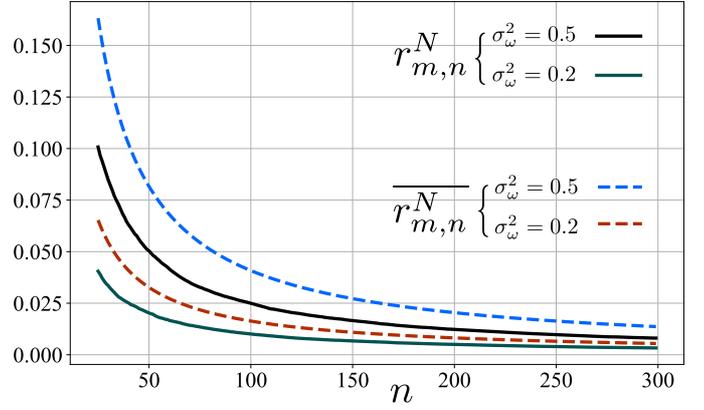}
\caption{ $r^N_{m, n}$ and estimated $\overline{r^N_{m, n}}$  with confidence and validation probabilities $P_\beta=95.45\%, \; =P_\alpha=95.45\%$ for $\sigma^2_\omega=0.2,\; 0.5$.}
\label{fig:rmnforn}
\end{figure}

Note that while the $\epsilon$-CoAC learnability in (\ref{eqn:CoacLearnable}) provides the desired minimum data length that is independent from the noise variance, the value of the dNMSE in (\ref{eqn:newlerSample}) is a function of the noise variance. Figure \ref{fig:rmnforn} shows the unavailable dNMSE (${r^{N}_{m, n}}$) and its estimated $\overline{r^{N}_{m, n}}$ in (\ref{eqn:upperrmnnew}) for two different values of noise variance, $\sigma^2_{\omega}=0.2$ and $\sigma^2_{\omega}=0.5$.


\begin{table}[]
\caption{The required data length $n$ for $\epsilon$-CoAC learnability based on the unavailable d2NMSE and its estimated probabilistic upperbound $\overline{r^{2N}_{m^*, n}}$ with confidence probability $P_\beta=95.45\%$ (known $m^*$) and with confidence and validation probabilities $ P_\beta=P_\alpha=95.45\%$ (unknown $m^*$) averaged over 1000 trials.}
\label{tab:nsampleTable2}
\resizebox{\columnwidth}{!}{\begin{tabular}{|c|c|c|c|c|}
\hline
$\epsilon$ &
  $n$(d2NMSE) &
  \begin{tabular} [c]{@{}c@{}} \Tstrut $n$ ( $\overline{r^{2N}_{m,n}}$)\\ ($m^*$ known)\Bstrut \end{tabular} &
  \begin{tabular}  [c]{@{}c@{}} \Tstrut  $n$ ($\overline{r^{2N}_{m,n}}$)\\ ($m^*$ unknown \\ with $\sigma^2_\omega = 0.2$) \Bstrut \end{tabular} &
  \begin{tabular}  [c]{@{}c@{}} \Tstrut $n$ ($\overline{r^{2N}_{m,n}}$)\\ ($m^*$ unknown \\ with $\sigma^2_\omega = 0.4$)  \Bstrut \end{tabular}  \\ \hline
0.1  & 24  & 43  & 62  & 89  \\ \hline
0.09 & 27  & 45  & 78  & 101 \\ \hline
0.08 & 30  & 51  & 83  & 142 \\ \hline
0.07 & 36  & 59  & 96  & 149 \\ \hline
0.06 & 42  & 68  & 108 & 164 \\ \hline
0.05 & 50  & 81  & 116 & 191 \\ \hline
0.04 & 63  & 104 & 139 & 226 \\ \hline
0.03 & 88  & 134 & 171 & 272 \\ \hline
0.02 & 131 & 202 & 270 & 304 \\ \hline
\end{tabular}}
\end{table}

\subsection{$\epsilon$-CoAC Learnability for when it is unknown if $\overline{\boldsymbol{\theta}}_{m^*} \in \mathcal{H}_m$.}
Here $\epsilon$-CoAC Learnability is examined for when a range of hypothesis classes $\mathcal{H}_m$ are compared and it is not known which one includes the unknown true parameter $m^*$ (Section \ref{section:HyCa}). 

\begin{figure}[htb]
\centering
\includegraphics[trim=0cm 0cm 0cm 0cm, scale=0.4]{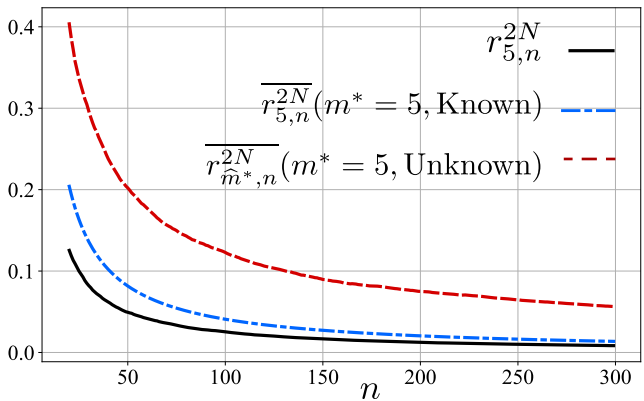}
\caption{Calculated $r^{2N}_{m, n}$ and $\overline{r^{2N}_{m, n}}$ with confidence probability $P_\beta=95.45\%$ for when the order complexity is known and with confidence and validation probabilities $P_\beta = P_\alpha = 95.45\%$ when the order complexity is estimated (not known) as a function of the length of the data ($\sigma^2_\omega=0.2$)}
\label{fig:differentms2}
\end{figure}

Figure \ref{fig:differentms2} shows the true $r^{2N}_{m,n}$ and $\overline{r^{2N}_{m,n}}$ , with confidence probability $ P_\beta=95.45\%$, for when $m^*$ is known and with confidence and validation probabilities $ P_\beta=P_\alpha=95.45\%$ when $m^*$ is unknown for two different noise variances. As expected, the upper bound becomes looser when $m^*$ is unknown and less information about the hyperparameter is available.

Table \ref{tab:nsampleTable2} also shows the minimum number of samples required in $\epsilon$-CoAC learnability for different values of $\epsilon$, averaged over 1000 trials. The columns shown under $n (\overline{r^{2N}_{m,n}})$ with unknown $m^*$ are the results of the probabilistic upper bound calculated with respect to (\ref{eqn:finalUpper}). The results are shown for two values of noise variance. Note that for the case of known $m^*$, the second column, the $\epsilon$-CoAC learnability in (\ref{eqn:generalNLearn}) is not a function of noise variance, however, in the case of unknown $m^*$, $\epsilon$-CoAC learnability is a function of noise variance (\ref{eqn:r2nwithrmsn}). As indicated in the table, the sample complexity required for the $\epsilon$-CoAC learnability increases for unknown $m^*$. Furthermore, as expected, a larger noise variance leads to a higher sample complexity, required for $\epsilon$-CoAC learnability.  

\begin{figure}[htb]
\centering
\includegraphics[trim=0cm 0cm 0cm 0cm, scale=0.41]{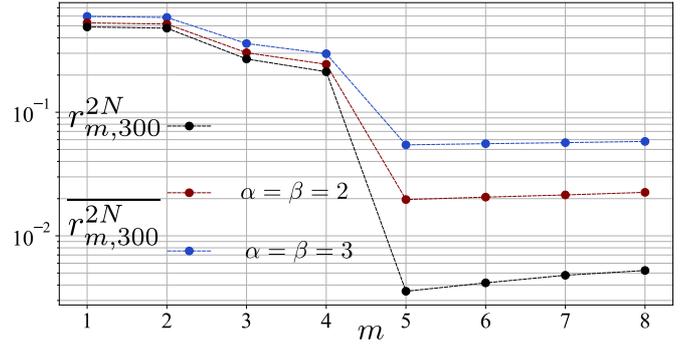}
\caption{Desired unavailable $r^{2N}_{m, n}$ and estimated $\overline{r^{2N}_{m, n}}$, $\sigma^2_\omega = 0.2$, for two different values of validation and confidence parameters, $\alpha$ and $\beta$.}
\label{fig:rmnalphabeta}
\end{figure}

So far, $\epsilon$ learnability has been examined for the required data length. Next, the result of the comparison of hypothesis classes with different orders (order selection (\ref{eqn:FullOptimization})) and the simultaneous calculation of $\epsilon_{min}$ (\ref{eqn:minepsilon}) are provided. Figure \ref{fig:rmnalphabeta} shows the desired $r^{2N}_{m, n}$  and $\overline{r^{2N}_{m, n}}$ for $\sigma^2_\omega = 0.2$, with two values of validation and confidence parameters, $\alpha$ and $\beta$ as a function of $m$ for when the data length is $n=300$. As the figure shows, both upper bounds are minimized at the true unknown $m^*=5$. 
Note that choosing a larger value for $\beta$ and $\alpha$, such as 3, that follows the three-sigma rule \cite{pukelsheim1994three}, will result in higher confidence and validation probabilities, $P_\beta=P_\alpha=99.7 \%$, while smaller $\beta$ and $\alpha$, such as 2, will result in a tighter bound with confidence and validation probabilities $P_\beta=P_\alpha=95 \%$. The choice of these probabilities is up to the learner. While the $\epsilon_{\min}$ value of 0.003 at $\widehat{m}^*=5$ indicates the minimum possible $\epsilon$ using the unavailable desired d2NMSE, using the $\epsilon$-CoAC learnability provides $\epsilon_{min}=0.024$ and $\epsilon_{min}=0.059$ for the considered two choices of confidence and validation probabilities, respectively.

Table \ref{tab:CV} compares the proposed hypothesis class selection method and the 5-fold cross-validation with respect to the average time cost and the average accuracy parameter (mean square error) MSE for two different data lengths of 300 and 100, for 1000 trials. Note that the computational complexity order of 5-fold cross-validation is $O(5Mn^3)$ while the computational complexity order of the proposed d2NMSE is $O(Mn^2)$ \cite{krstajic2014cross}.
As the table indicates, not only does the proposed method require much less computation time, but it also outperforms the cross-validation method in the sense of the average of the MSE error.  
\begin{table*}[htb]
\centering
\caption{Time cost and accuracy of the proposed d2NMSE with $P_\beta=99.7,\%=P_\alpha=99.7\%$ and 5-fold cross-validation (CV), for two different data length $n=100$, $n=300$}
\label{tab:CV}
\begin{tabular}{|c|c|cc||cc||cc||cc|}
\hline
\multirow{2}{*}{$\sigma^2_\omega$} & \multirow{2}{*}{SNR-dB} & \multicolumn{2}{c|}{\begin{tabular}[c]{@{}c@{}}Time (m:s.ms)\\ n=300\end{tabular}} & \multicolumn{2}{c|}{\begin{tabular}[c]{@{}c@{}}Accuracy(MSE)\\ n=300\end{tabular}} & \multicolumn{2}{c|}{\begin{tabular}[c]{@{}c@{}}Time (m:s.ms)\\ n=100\end{tabular}} & \multicolumn{2}{c|}{\begin{tabular}[c]{@{}c@{}}Accuracy(MSE)\\ n=100\end{tabular}} \\ \cline{3-10} 
 &  & \multicolumn{1}{c|}{\begin{tabular}[c]{@{}c@{}}proposed\\ d2NMSE\end{tabular}} & \begin{tabular}[c]{@{}c@{}}5-fold\\ CV\end{tabular} & \multicolumn{1}{c|}{\begin{tabular}[c]{@{}c@{}}proposed\\ d2NMSE\end{tabular}} & \begin{tabular}[c]{@{}c@{}}5-fold\\ CV\end{tabular} & \multicolumn{1}{c|}{\begin{tabular}[c]{@{}c@{}}proposed\\ d2NMSE\end{tabular}} & \begin{tabular}[c]{@{}c@{}}5-fold\\ CV\end{tabular} & \multicolumn{1}{c|}{\begin{tabular}[c]{@{}c@{}}proposed\\ d2NMSE\end{tabular}} & \begin{tabular}[c]{@{}c@{}}5-fold\\ CV\end{tabular} \\ \hline
0.9 & -2.513 & \multicolumn{1}{c|}{\textbf{0:1.7}} & 6:47.6 & \multicolumn{1}{c|}{\textbf{0.074}} & 0.87 & \multicolumn{1}{c|}{\textbf{0:1.1}} & 5:26.1 & \multicolumn{1}{c|}{\textbf{0.212}} & 0.412 \\ \hline
0.8 & -2.002 & \multicolumn{1}{c|}{\textbf{0:1.6}} & 6:38.1 & \multicolumn{1}{c|}{\textbf{0.061}} & 0.071 & \multicolumn{1}{c|}{\textbf{0:1.2}} & 5:53.0 & \multicolumn{1}{c|}{\textbf{0.178}} & 0.311 \\ \hline
0.7 & -1.422 & \multicolumn{1}{c|}{\textbf{0:1.7}} & 6:52.2 & \multicolumn{1}{c|}{\textbf{0.046}} & 0.046 & \multicolumn{1}{c|}{\textbf{0:1.1}} & 5:17.1 & \multicolumn{1}{c|}{\textbf{0.167}} & 0.282 \\ \hline
0.6 & -0.753 & \multicolumn{1}{c|}{\textbf{0:1.6}} & 6:51.7 & \multicolumn{1}{c|}{\textbf{0.038}} & 0.039 & \multicolumn{1}{c|}{\textbf{0:1.3}} & 5:20.3 & \multicolumn{1}{c|}{\textbf{0.139}} & 0.267 \\ \hline
0.5 & 0.038 & \multicolumn{1}{c|}{\textbf{0:1.5}} & 6:48.4 & \multicolumn{1}{c|}{\textbf{0.031}} & 0.032 & \multicolumn{1}{c|}{\textbf{0:1.2}} & 5:31.9 & \multicolumn{1}{c|}{\textbf{0.119}} & 0.265 \\ \hline
0.4 & 1.007 & \multicolumn{1}{c|}{\textbf{0:1.8}} & 6:43.2 & \multicolumn{1}{c|}{\textbf{0.019}} & \textbf{0.019} & \multicolumn{1}{c|}{\textbf{0:1.2}} & 5:22.1 & \multicolumn{1}{c|}{\textbf{0.097}} & 0.214 \\ \hline
0.3 & 2.257 & \multicolumn{1}{c|}{\textbf{0:1:6}} & 6:50.4 & \multicolumn{1}{c|}{\textbf{0.014}} & \textbf{0.014} & \multicolumn{1}{c|}{\textbf{0:1.1}} & 5:10.4 & \multicolumn{1}{c|}{\textbf{0.074}} & 0.186 \\ \hline
0.2 & 4.018 & \multicolumn{1}{c|}{\textbf{0:1.7}} & 6:47.3 & \multicolumn{1}{c|}{\textbf{0.00}} & 0.01 & \multicolumn{1}{c|}{\textbf{0:1.3}} & 5:12.6 & \multicolumn{1}{c|}{\textbf{0.047}} & 0.141 \\ \hline
0.1 & 7.028 & \multicolumn{1}{c|}{\textbf{0:1.5}} & 6:49.6 & \multicolumn{1}{c|}{\textbf{0.00}} & \textbf{0.00} & \multicolumn{1}{c|}{\textbf{0:1.2}} & 5:32.1 & \multicolumn{1}{c|}{\textbf{0.024}} & 0.095 \\ \hline
\end{tabular}
\end{table*}
It is important to mention that when the data length are 100 or 300, the estimated $m^*$ is consistently the true value 5 for both methods that are compared in Table \ref{tab:CV}. However, as the data length decreases, the 5-fold validation algorithm starts failing to choose the correct number of parameters. Table \ref{tab:esm} compares the methods with respect to MSE, as well as the average and standard deviation of the estimated parameter complexity of the hypothesis class, when the data length is $n=50$, with confidence and validation parameters $\alpha = \beta =3$ (averaged for 1000 runs). As the table shows, the proposed method in this case outperforms the 5-fold cross-validation not only in the MSE sense, but also in the sense of finding the true number of parameters of the regression. If the goal is to find the order of the polynomial in the regression, as the table shows, the 5-fold cross-validation approach fails for a range of noise variances and starts to get close to the true $m^*$ for very small noise variances. This is due to the fact that the k-fold cross-validation method loses a significant portion of the data set as a holdout validation set and, therefore, fails to estimate the parameter vector accurately, while the proposed method benefits from using all of the training data set, without splitting, in the model estimation process. 

\begin{table}[htb]
\footnotesize
\centering
\caption{The estimated optimal order $\widehat{m}^*$ for data length $n=50$ with $P_\beta=99.7,\%=P_\alpha=99.7\%$}
\label{tab:esm}
\resizebox{\columnwidth}{!}{\begin{tabular}{|c|c|cc||cc||cc|}
\hline
\multirow{2}{*}{$\sigma^2_\omega$} & \multirow{2}{*}{SNR} & \multicolumn{2}{c|}{\begin{tabular}[c]{@{}c@{}}Average\\ MSE\end{tabular}} & \multicolumn{2}{c|}{\begin{tabular}[c]{@{}c@{}}average\\ $\widehat{m}$\end{tabular}} & \multicolumn{2}{c|}{\begin{tabular}[c]{@{}c@{}}SD\\ $\widehat{m}$\end{tabular}} \\ \cline{3-8} 
 &  & \multicolumn{1}{c|}{\begin{tabular}[c]{@{}c@{}}Proposed\\ d2NMSE\end{tabular}} & \begin{tabular}[c]{@{}c@{}}5-fold\\ CV\end{tabular} & \multicolumn{1}{c|}{\begin{tabular}[c]{@{}c@{}}Proposed\\ d2NMSE\end{tabular}} & \begin{tabular}[c]{@{}c@{}}5-fold\\ CV\end{tabular} & \multicolumn{1}{c|}{\begin{tabular}[c]{@{}c@{}}Proposed\\ d2NMSE\end{tabular}} & \begin{tabular}[c]{@{}c@{}}5-fold\\ CV\end{tabular} \\ \hline
0.9 & -2.513 & \multicolumn{1}{c|}{\textbf{0.84}} & 1.13 & \multicolumn{1}{c|}{\textbf{4.98}} & 3.1 & \multicolumn{1}{c|}{\textbf{1.01}} & 1.51 \\ \hline
0.8 & -2.002 & \multicolumn{1}{c|}{\textbf{0.83}} & 1.03 & \multicolumn{1}{c|}{\textbf{5.01}} & 3.25 & \multicolumn{1}{c|}{\textbf{0.98}} & 1.04 \\ \hline
0.7 & -1.422 & \multicolumn{1}{c|}{\textbf{0.77}} & 0.96 & \multicolumn{1}{c|}{\textbf{5.10}} & 3.38 & \multicolumn{1}{c|}{\textbf{0.87}} & 1.01 \\ \hline
0.6 & -0.753 & \multicolumn{1}{c|}{\textbf{0.71}} & 0.87 & \multicolumn{1}{c|}{\textbf{5.07}} & 3.42 & \multicolumn{1}{c|}{\textbf{0.76}} & 0.96 \\ \hline
0.5 & 0.038 & \multicolumn{1}{c|}{\textbf{0.61}} & 0.75 & \multicolumn{1}{c|}{\textbf{5.03}} & 3.55 & \multicolumn{1}{c|}{\textbf{0.65}} & 0.91 \\ \hline
0.4 & 1.007 & \multicolumn{1}{c|}{\textbf{0.31}} & 0.55 & \multicolumn{1}{c|}{\textbf{5.14}} & 3.89 & \multicolumn{1}{c|}{\textbf{0.52}} & 0.88 \\ \hline
0.3 & 2.257 & \multicolumn{1}{c|}{\textbf{0.21}} & 0.26 & \multicolumn{1}{c|}{\textbf{5.05}} & 4.01 & \multicolumn{1}{c|}{\textbf{0.14}} & 0.86 \\ \hline
0.2 & 4.018 & \multicolumn{1}{c|}{\textbf{0.7}} & 0.14 & \multicolumn{1}{c|}{\textbf{5.01}} & 4.19 & \multicolumn{1}{c|}{\textbf{0.06}} & 0.72 \\ \hline
0.1 & 7.028 & \multicolumn{1}{c|}{\textbf{0.00}} & 0.02 & \multicolumn{1}{c|}{\textbf{5.07}} & 4.55 & \multicolumn{1}{c|}{\textbf{0.05}} & 0.61 \\ \hline
\end{tabular}}
\end{table}

\section{Conclusion}

The proposed $\epsilon$-Confidence Approximately Correct ($\epsilon$-CoAC)  takes the uncertainty in the target vector into account in defining new foundations for hypothesis class learnability. The approach employs the Kullback-Leibler divergence to define Typical set for hypotheses that are approximately correct with respect to the desired target vector. The theory and application of this framework are shown for the regression model and are analyzed with respect to the length of the training data set and hypothesis class complexity order. 

The proposed method is capable of estimating the required sample complexity for an accurate learner $\epsilon$, but it can also compare hypothesis classes with different complexity orders. The approach provides a clear connection of learnability with training sample size, hypothesis class complexity, as well as the relative power of uncertainty and target vector. The simulation results show superiority of the proposed learning algorithm to the well-known cross-validation method for determining the optimal hypothesis class, in terms of accuracy and computation time. The $\epsilon$-CoAC learnability is a promising approach that can be extended for the learning process of piecewise linear models and for application in the learnability, validation, and structural analysis and comparison of multi layer perceptron (MLP).


\bibliographystyle{elsarticle-num} 
\footnotesize
\bibliography{mybibfile}

\begin{thebibliography}{10}
\expandafter\ifx\csname url\endcsname\relax
  \def\url#1{\texttt{#1}}\fi
\expandafter\ifx\csname urlprefix\endcsname\relax\def\urlprefix{URL }\fi
\expandafter\ifx\csname href\endcsname\relax
  \def\href#1#2{#2} \def\path#1{#1}\fi

\bibitem{vapnik1999overview}
V.~N. Vapnik, An overview of statistical learning theory, IEEE transactions on
  neural networks 10~(5) (1999) 988--999.

\bibitem{evgeniou2002regularization}
T.~Evgeniou, T.~Poggio, M.~Pontil, A.~Verri, Regularization and statistical
  learning theory for data analysis, Computational Statistics \& Data Analysis
  38~(4) (2002) 421--432.

\bibitem{shailaja2018machine}
K.~Shailaja, B.~Seetharamulu, M.~Jabbar, Machine learning in healthcare: A
  review, in: 2018 Second international conference on electronics,
  communication and aerospace technology (ICECA), IEEE, 2018, pp. 910--914.

\bibitem{qayyum2020secure}
A.~Qayyum, J.~Qadir, M.~Bilal, A.~Al-Fuqaha, Secure and robust machine learning
  for healthcare: A survey, IEEE Reviews in Biomedical Engineering 14 (2020)
  156--180.

\bibitem{wiens2018machine}
J.~Wiens, E.~S. Shenoy, Machine learning for healthcare: on the verge of a
  major shift in healthcare epidemiology, Clinical Infectious Diseases 66~(1)
  (2018) 149--153.

\bibitem{culkin2017machine}
R.~Culkin, S.~R. Das, Machine learning in finance: the case of deep learning
  for option pricing, Journal of Investment Management 15~(4) (2017) 92--100.

\bibitem{dixon2020machine}
M.~F. Dixon, I.~Halperin, P.~Bilokon, Machine learning in Finance, Vol. 1406,
  Springer, 2020.

\bibitem{tarca2007machine}
A.~L. Tarca, V.~J. Carey, X.-w. Chen, R.~Romero, S.~Dr{\u{a}}ghici, Machine
  learning and its applications to biology, PLoS computational biology 3~(6)
  (2007) e116.

\bibitem{oliveira2019biotechnology}
A.~L. Oliveira, Biotechnology, big data and artificial intelligence,
  Biotechnology journal 14~(8) (2019) 1800613.

\bibitem{xin2018machine}
Y.~Xin, L.~Kong, Z.~Liu, Y.~Chen, Y.~Li, H.~Zhu, M.~Gao, H.~Hou, C.~Wang,
  Machine learning and deep learning methods for cybersecurity, Ieee access 6
  (2018) 35365--35381.

\bibitem{handa2019machine}
A.~Handa, A.~Sharma, S.~K. Shukla, Machine learning in cybersecurity: A review,
  Wiley Interdisciplinary Reviews: Data Mining and Knowledge Discovery 9~(4)
  (2019) e1306.

\bibitem{ghahramani2015probabilistic}
Z.~Ghahramani, Probabilistic machine learning and artificial intelligence,
  Nature 521~(7553) (2015) 452--459.

\bibitem{vapnik1999nature}
V.~Vapnik, The nature of statistical learning theory, Springer science \&
  business media, 1999.

\bibitem{wolpert1997no}
D.~H. Wolpert, W.~G. Macready, No free lunch theorems for optimization, IEEE
  transactions on evolutionary computation 1~(1) (1997) 67--82.

\bibitem{adam2019no}
S.~P. Adam, S.-A.~N. Alexandropoulos, P.~M. Pardalos, M.~N. Vrahatis, No free
  lunch theorem: A review, Approximation and optimization (2019) 57--82.

\bibitem{lugosi2002pattern}
G.~Lugosi, Pattern classification and learning theory, in: Principles of
  nonparametric learning, Springer, 2002, pp. 1--56.

\bibitem{shalev2014understanding}
S.~Shalev-Shwartz, S.~Ben-David, Understanding machine learning: From theory to
  algorithms, Cambridge university press, 2014.

\bibitem{hastie2009elements}
T.~Hastie, R.~Tibshirani, J.~H. Friedman, J.~H. Friedman, The elements of
  statistical learning: data mining, inference, and prediction, Vol.~2,
  Springer, 2009.

\bibitem{valiant1984theory}
L.~G. Valiant, A theory of the learnable, Communications of the ACM 27~(11)
  (1984) 1134--1142.

\bibitem{arlot2010survey}
S.~Arlot, A.~Celisse, A survey of cross-validation procedures for model
  selection (2010).

\bibitem{cover1999elements}
T.~M. Cover, Elements of information theory, John Wiley \& Sons, 1999.

\bibitem{long1995sample}
P.~M. Long, On the sample complexity of pac learning half-spaces against the
  uniform distribution, IEEE Transactions on Neural Networks 6~(6) (1995)
  1556--1559.

\bibitem{golowich2018size}
N.~Golowich, A.~Rakhlin, O.~Shamir, Size-independent sample complexity of
  neural networks, in: Conference On Learning Theory, PMLR, 2018, pp. 297--299.

\bibitem{wang2022generalizing}
J.~Wang, C.~Lan, C.~Liu, Y.~Ouyang, T.~Qin, W.~Lu, Y.~Chen, W.~Zeng, P.~Yu,
  Generalizing to unseen domains: A survey on domain generalization, IEEE
  Transactions on Knowledge and Data Engineering (2022).

\bibitem{farahani2021brief}
A.~Farahani, S.~Voghoei, K.~Rasheed, H.~R. Arabnia, A brief review of domain
  adaptation, Advances in Data Science and Information Engineering: Proceedings
  from ICDATA 2020 and IKE 2020 (2021) 877--894.

\bibitem{gretton2009covariate}
A.~Gretton, A.~Smola, J.~Huang, M.~Schmittfull, K.~Borgwardt, B.~Sch{\"o}lkopf,
  Covariate shift by kernel mean matching, Dataset shift in machine learning
  3~(4) (2009) 5.

\bibitem{wen2014robust}
J.~Wen, C.-N. Yu, R.~Greiner, Robust learning under uncertain test
  distributions: Relating covariate shift to model misspecification, in:
  International Conference on Machine Learning, PMLR, 2014, pp. 631--639.

\bibitem{zhou2022domain}
K.~Zhou, Z.~Liu, Y.~Qiao, T.~Xiang, C.~C. Loy, Domain generalization: A survey,
  IEEE Transactions on Pattern Analysis and Machine Intelligence (2022).

\bibitem{ghifary2016scatter}
M.~Ghifary, D.~Balduzzi, W.~B. Kleijn, M.~Zhang, Scatter component analysis: A
  unified framework for domain adaptation and domain generalization, IEEE
  transactions on pattern analysis and machine intelligence 39~(7) (2016)
  1414--1430.

\bibitem{mahajan2021domain}
D.~Mahajan, S.~Tople, A.~Sharma, Domain generalization using causal matching,
  in: International Conference on Machine Learning, PMLR, 2021, pp. 7313--7324.

\bibitem{mehrabi2021survey}
N.~Mehrabi, F.~Morstatter, N.~Saxena, K.~Lerman, A.~Galstyan, A survey on bias
  and fairness in machine learning, ACM Computing Surveys (CSUR) 54~(6) (2021)
  1--35.

\bibitem{zafar2019fairness}
M.~B. Zafar, I.~Valera, M.~Gomez-Rodriguez, K.~P. Gummadi, Fairness
  constraints: A flexible approach for fair classification, The Journal of
  Machine Learning Research 20~(1) (2019) 2737--2778.

\bibitem{chamon2020empirical}
L.~F. Chamon, S.~Paternain, M.~Calvo-Fullana, A.~Ribeiro, The empirical duality
  gap of constrained statistical learning, in: ICASSP 2020-2020 IEEE
  International Conference on Acoustics, Speech and Signal Processing (ICASSP),
  IEEE, 2020, pp. 8374--8378.

\bibitem{song2022learning}
H.~Song, M.~Kim, D.~Park, Y.~Shin, J.-G. Lee, Learning from noisy labels with
  deep neural networks: A survey, IEEE Transactions on Neural Networks and
  Learning Systems (2022).

\bibitem{ma2018dimensionality}
X.~Ma, Y.~Wang, M.~E. Houle, S.~Zhou, S.~Erfani, S.~Xia, S.~Wijewickrema,
  J.~Bailey, Dimensionality-driven learning with noisy labels, in:
  International Conference on Machine Learning, PMLR, 2018, pp. 3355--3364.

\bibitem{zhang2021learning}
M.~Zhang, J.~Lee, S.~Agarwal, Learning from noisy labels with no change to the
  training process, in: International Conference on Machine Learning, PMLR,
  2021, pp. 12468--12478.

\bibitem{han2018masking}
B.~Han, J.~Yao, G.~Niu, M.~Zhou, I.~Tsang, Y.~Zhang, M.~Sugiyama, Masking: A
  new perspective of noisy supervision, Advances in neural information
  processing systems 31 (2018).

\bibitem{wang2019symmetric}
Y.~Wang, X.~Ma, Z.~Chen, Y.~Luo, J.~Yi, J.~Bailey, Symmetric cross entropy for
  robust learning with noisy labels, in: Proceedings of the IEEE/CVF
  International Conference on Computer Vision, 2019, pp. 322--330.

\bibitem{wong2016constrained}
S.~Y. Wong, K.~S. Yap, H.~J. Yap, A constrained optimization based extreme
  learning machine for noisy data regression, Neurocomputing 171 (2016)
  1431--1443.

\bibitem{kim2009estimating}
J.-H. Kim, Estimating classification error rate: Repeated cross-validation,
  repeated hold-out and bootstrap, Computational statistics \& data analysis
  53~(11) (2009) 3735--3745.

\bibitem{cherkassky2003comparison}
V.~Cherkassky, Y.~Ma, Comparison of model selection for regression, Neural
  computation 15~(7) (2003) 1691--1714.

\bibitem{fushiki2011estimation}
T.~Fushiki, Estimation of prediction error by using k-fold cross-validation,
  Statistics and Computing 21 (2011) 137--146.

\bibitem{ranstam2018lasso}
J.~Ranstam, J.~Cook, Lasso regression, Journal of British Surgery 105~(10)
  (2018) 1348--1348.

\bibitem{marquardt1975ridge}
D.~W. Marquardt, R.~D. Snee, Ridge regression in practice, The American
  Statistician 29~(1) (1975) 3--20.

\bibitem{friedman2010regularization}
J.~Friedman, T.~Hastie, R.~Tibshirani, Regularization paths for generalized
  linear models via coordinate descent, Journal of statistical software 33~(1)
  (2010) 1.

\bibitem{awad2015support}
M.~Awad, R.~Khanna, M.~Awad, R.~Khanna, Support vector regression, Efficient
  learning machines: Theories, concepts, and applications for engineers and
  system designers (2015) 67--80.

\bibitem{balasundaram2020robust}
S.~Balasundaram, S.~C. Prasad, Robust twin support vector regression based on
  huber loss function, Neural Computing and Applications 32 (2020)
  11285--11309.

\bibitem{domingos2000bayesian}
P.~Domingos, Bayesian averaging of classifiers and the overfitting problem, in:
  ICML, Vol. 747, 2000, pp. 223--230.

\bibitem{myung2006model}
J.~I. Myung, D.~J. Navarro, M.~A. Pitt, Model selection by normalized maximum
  likelihood, Journal of Mathematical Psychology 50~(2) (2006) 167--179.

\bibitem{patel1996handbook}
J.~K. Patel, C.~B. Read, Handbook of the normal distribution, Vol. 150, CRC
  Press, 1996.

\bibitem{guo2011estimation}
D.~Guo, Y.~Wu, S.~S. Shitz, S.~Verd{\'u}, Estimation in gaussian noise:
  Properties of the minimum mean-square error, IEEE Transactions on Information
  Theory 57~(4) (2011) 2371--2385.

\bibitem{montgomery2021introduction}
D.~C. Montgomery, E.~A. Peck, G.~G. Vining, Introduction to linear regression
  analysis, John Wiley \& Sons, 2021.

\bibitem{beheshti2009noisy}
S.~Beheshti, M.~A. Dahleh, Noisy data and impulse response estimation, IEEE
  Transactions on Signal Processing 58~(2) (2009) 510--521.

\bibitem{shamsi2022relative}
M.~Shamsi, S.~Beheshti, Relative entropy (re) based lti system modeling
  equipped with time delay estimation and online modeling, arXiv preprint
  arXiv:2210.01279 (2022).

\bibitem{pukelsheim1994three}
F.~Pukelsheim, The three sigma rule, The American Statistician 48~(2) (1994)
  88--91.

\bibitem{krstajic2014cross}
D.~Krstajic, L.~J. Buturovic, D.~E. Leahy, S.~Thomas, Cross-validation pitfalls
  when selecting and assessing regression and classification models, Journal of
  cheminformatics 6 (2014) 1--15.

\end{thebibliography}





\end{document}